%% file: main.tex
\documentclass{article}

\usepackage{arxiv}

\usepackage[utf8]{inputenc}
\usepackage[T1]{fontenc}
\usepackage{lmodern}
\usepackage{microtype}

\usepackage{amsmath}
\usepackage{amsfonts}
\usepackage{amssymb}
\usepackage{mathtools}
\usepackage{amsthm}
\usepackage{bbold}
\usepackage{bm}
\usepackage{nicefrac}

\usepackage{xcolor}
\usepackage{graphicx}
\graphicspath{{./images/}{./Figures/}{./figures/}}

\usepackage{booktabs}
\usepackage{threeparttable}
\usepackage{tabularray}
\usepackage{tabularx}
\usepackage{adjustbox}
\usepackage{array}
\usepackage{makecell}
\usepackage{float}
\usepackage{subfigure}
\usepackage{caption}
\usepackage{colortbl}
\usepackage{etoolbox}

\usepackage{algorithm}
\usepackage[algo2e]{algorithm2e}

\usepackage{algorithmic}

\usepackage[most]{tcolorbox}
\usepackage{enumitem}
\usepackage{pifont}
\usepackage{fvextra}

\IfFileExists{fontawesome5.sty}{
  \usepackage{fontawesome5}
  \newcommand{\githubicon}{\faGithub}
}{
  \newcommand{\githubicon}{\textsc{GitHub}}
}

\usepackage[numbers]{natbib}
\usepackage{url}

\definecolor{sghaBlue}{HTML}{0B4A8B}
\definecolor{sghaLightBlue}{HTML}{EAF3FF}
\definecolor{sghaBorderBlue}{HTML}{78AEE8}
\definecolor{sghaTeal}{HTML}{0B7A75}
\definecolor{sghaOrange}{HTML}{C75B12}
\definecolor{sghaPurple}{HTML}{6A3DA8}
\definecolor{sghaGray}{HTML}{F6F7F9}
\definecolor{sghaDarkGray}{HTML}{333333}
\definecolor{sghaCite}{HTML}{B0177A}
\definecolor{sghaTableHeader}{HTML}{0B3A6E}
\definecolor{sghaTotalRow}{HTML}{EEF3F8}

\usepackage[
  colorlinks=true,
  linkcolor=sghaBlue,
  citecolor=sghaCite,
  urlcolor=sghaBlue
]{hyperref}
\usepackage[nameinlink,capitalise,noabbrev]{cleveref}

\usepackage{titlesec}

\titleformat{\section}
  {\Large\bfseries\color{sghaBlue}}
  {\thesection}
  {0.7em}
  {}

\titleformat{\subsection}
  {\large\bfseries\color{sghaBlue}}
  {\thesubsection}
  {0.7em}
  {}

\titleformat{\subsubsection}
  {\normalsize\bfseries\color{sghaDarkGray}}
  {\thesubsubsection}
  {0.7em}
  {}

\setlist[itemize]{leftmargin=1.4em,itemsep=0.25em,topsep=0.25em}
\setlist[enumerate]{leftmargin=1.6em,itemsep=0.25em,topsep=0.25em}

\newcolumntype{L}[1]{>{\raggedright\arraybackslash}p{#1}}
\newcolumntype{C}[1]{>{\centering\arraybackslash}p{#1}}
\newcolumntype{R}[1]{>{\raggedleft\arraybackslash}p{#1}}

\newcommand{\tableheader}{\rowcolor{sghaTableHeader}}
\newcommand{\tabtotal}{\rowcolor{sghaTotalRow}}
\newcommand{\theadcell}[1]{\textcolor{white}{\textbf{\makecell{#1}}}}
\newcommand{\bestscore}[1]{\textbf{\textcolor{sghaBlue}{#1}}}

\newcommand{\cmark}{\ding{51}}
\newcommand{\xmark}{\ding{55}}
\newcommand{\yesfield}{\textcolor{sghaTeal}{\cmark}}
\newcommand{\nofield}{\textcolor{sghaOrange}{\xmark}}

\AtBeginEnvironment{table}{%
  \centering
  \small
  \setlength{\tabcolsep}{4pt}%
  \renewcommand{\arraystretch}{1.12}%
}

\AtBeginEnvironment{table*}{%
  \centering
  \small
  \setlength{\tabcolsep}{4pt}%
  \renewcommand{\arraystretch}{1.12}%
}

\BeforeBeginEnvironment{tabular}{\begin{adjustbox}{max width=\linewidth,center}}
\AfterEndEnvironment{tabular}{\end{adjustbox}}

\newtcolorbox{contributionbox}[1][Contributions]{
  enhanced,
  breakable,
  colback=sghaLightBlue,
  colframe=sghaBorderBlue,
  colbacktitle=sghaBlue,
  coltitle=white,
  title={#1},
  fonttitle=\bfseries,
  boxrule=0.7pt,
  arc=2pt,
  left=8pt,
  right=8pt,
  top=8pt,
  bottom=8pt,
  before skip=1.0em,
  after skip=1.0em
}

\newtcolorbox{sghainfobox}[1]{
  enhanced,
  breakable,
  colback=sghaLightBlue,
  colframe=sghaBorderBlue,
  colbacktitle=sghaBlue,
  coltitle=white,
  title={#1},
  fonttitle=\bfseries,
  boxrule=0.6pt,
  arc=2pt,
  left=8pt,
  right=8pt,
  top=7pt,
  bottom=7pt,
  before skip=0.9em,
  after skip=0.9em
}

\newtcolorbox{relatedworkbox}[1]{
  enhanced,
  breakable,
  colback=white,
  colframe=sghaBorderBlue,
  colbacktitle=sghaLightBlue,
  coltitle=sghaBlue,
  title={#1},
  fonttitle=\bfseries,
  boxrule=0.55pt,
  arc=2pt,
  left=8pt,
  right=8pt,
  top=7pt,
  bottom=7pt,
  before skip=0.9em,
  after skip=0.9em
}

\newtcolorbox{artifactbox}[1]{
  enhanced,
  breakable,
  colback=white,
  colframe=sghaBorderBlue,
  colbacktitle=sghaTableHeader,
  coltitle=white,
  title={#1},
  fonttitle=\bfseries\small,
  boxrule=0.65pt,
  arc=2pt,
  left=8pt,
  right=8pt,
  top=8pt,
  bottom=8pt,
  before skip=0.9em,
  after skip=0.9em
}

\newcommand{\artifactfield}[1]{\vspace{0.45em}\noindent\textcolor{sghaBlue}{\textbf{#1.}}}

\DefineVerbatimEnvironment{PromptBox}{Verbatim}{
  breaklines=true,
  breakanywhere=true,
  fontsize=\footnotesize
}

\theoremstyle{plain}

\theoremstyle{definition}

\theoremstyle{remark}

\newcommand{\ignore}[1]{}
\allowdisplaybreaks[4]

\newcommand{\authorlogoh}{1.05em}

\DeclareRobustCommand{\IITBMark}{%
  \IfFileExists{Figures/iitb.png}
  {\raisebox{-0.18em}{\includegraphics[height=\authorlogoh,keepaspectratio]{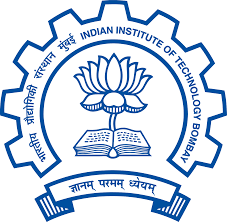}}}
  {\textsuperscript{\textsc{IITB}}}%
}

\DeclareRobustCommand{\MBZUAIMark}{%
  \IfFileExists{Figures/mbzuai.jpg}
  {\raisebox{-0.18em}{\includegraphics[height=\authorlogoh,keepaspectratio]{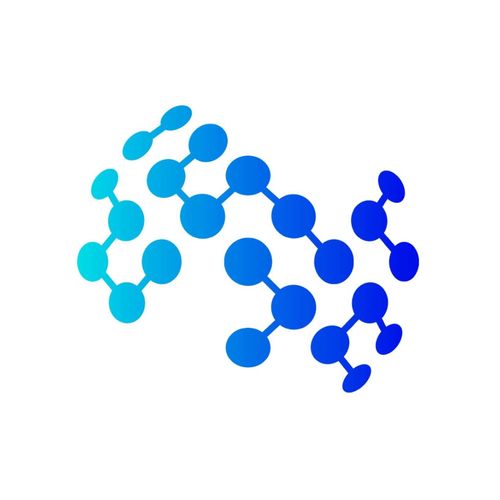}}}
  {\textsuperscript{\textsc{MBZUAI}}}%
}

\title{SGHA: Evidence-Grounded Research Problem Discovery with Local Language Models}

\author{
\begin{tabular}{c}
\textbf{Sarvesh Gharat}\,\IITBMark
\qquad
\textbf{Junpei Komiyama}\,\MBZUAIMark
\\[0.45em]
{\small \IITBMark\; C-MInDS, IIT Bombay}
\qquad
{\small \MBZUAIMark\; Machine Learning Department, MBZUAI}
\\[0.35em]
{\small
\href{mailto:sarveshgharat19@gmail.com}{\texttt{sarveshgharat19@gmail.com}}
\qquad
\href{mailto:junpei.komiyama@gmail.com}{\texttt{junpei.komiyama@gmail.com}}
}
\end{tabular}
}

\makeatletter
\renewenvironment{abstract}{
  \vskip 0.075in
  \centerline{\large\bfseries\color{sghaBlue} Abstract}
  \vspace{0.5ex}
  \begin{quote}
}{
  \par\end{quote}\vskip 1ex
}
\makeatother
\begin{document}

\maketitle

\begin{abstract}
Recent efforts toward fully automated AI scientists have demonstrated that language-model agents can generate hypotheses, execute experiments, and draft scientific manuscripts. However, during the early stages of research, when research problems are formulated, these AI scientists often rely heavily on proprietary frontier models. Their proposals are shaped by opaque parametric knowledge and by literature searches conditioned on the proposals themselves. Such knowledge is effectively a black box, and this dependence makes the evidential basis and validity of generated research problems difficult to audit and leaves the process vulnerable to model-specific hallucinations and biases. Furthermore, if proprietary research materials are transmitted to external APIs, the use of these models creates confidentiality, privacy, and data-governance concerns.

We introduce the Structural Gap Hypothesis Agent (SGHA), a fully automated, corpus-first research-problem discovery system that runs entirely on a local LLM. SGHA structures a scientific literature corpus into evidence-linked paper objects and a typed evidence graph, detects unresolved structural patterns across papers, screens candidate gaps before formulation, and produces traceable research-problem families. In particular, it is able to output assumptions, objectives, success criteria, and remaining ambiguities. All LLM-based components of SGHA are executed using a locally served open-weight 9B language model, without requiring proprietary frontier-model APIs.
We compare SGHA with the AI Scientist-v2 idea formulation module in five machine-learning domains. Our results suggest that explicit corpus structure and evidence-constrained reasoning can support promising, inspectable research-problem formulation without relying on frontier models during generation or verification.
\end{abstract}

\vspace{0.5em}
\begin{center}
\small
\href{https://github.com/SarveshVGharat/structural-gap-hypothesis-agent}
{\githubicon\ \texttt{github.com/SarveshVGharat/structural-gap-hypothesis-agent}}
\end{center}
\vspace{0.5em}

\input{files/introduction}
\input{files/sgha}
\input{files/experimental_setup}
\input{files/results}
\input{files/discussion}
\input{files/conclusion}
\input{files/limitation}
\input{files/genai_disclosure}

\newpage

\bibliographystyle{unsrtnat}
\bibliography{main}

\newpage

\input{appendix/appendixa}
\input{appendix/appendixb}
\input{appendix/appendixc}

\end{document}

%% file: files/introduction.tex
\section{Introduction}

Recent language-model agents can generate research ideas and automate parts of the scientific workflow, including writing code, running experiments, analyzing results, and preparing manuscripts~\citep{zhou2024hypothesis, lu2024ai, yamada2025ai, zhang2024comprehensive, gottweis2025towards, ren2025towards, baek2025researchagent}. Several end-to-end systems begin by generating candidate ideas within a user-provided topic or experimental scaffold and then try to develop them. In this paper, we study a different starting point. We ask whether an automated system can identify research problems that are already suggested by unresolved patterns in the scientific literature, even when no single paper states them directly.

\begin{figure*}[!t]
\centering
\includegraphics[width=0.96\textwidth]{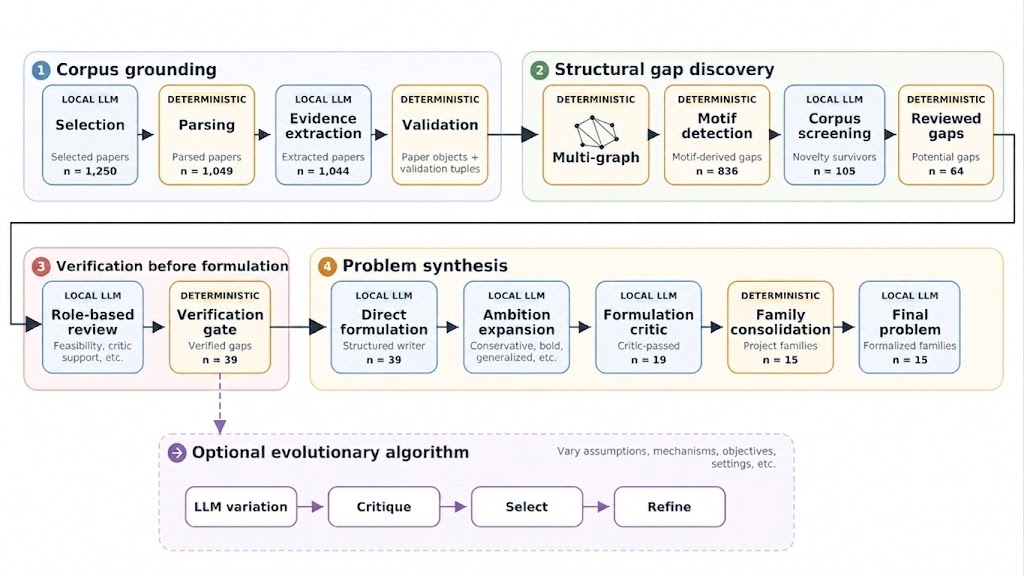}
\caption{Overview of the SGHA pipeline. ``Deterministic'' denotes non-LLM stages. The evolutionary branch is optional and is reported separately from the verification-gated family outputs. Evolutionary operations are iterated multiple times and fed into the problem synthesis stage.}
\label{fig:sgha-overview}
\end{figure*}

Many research questions only become visible when several papers are considered together. A set of methods may fail under the same condition, several guarantees may rely on the same strong assumption, or a limitation may be reported repeatedly without being turned into a concrete research problem. In such cases, the evidence for a problem is already present, but it is distributed across papers.

As a simple example, consider the literature on stochastic multi-armed bandits from 1985 through 2011.
Researchers try to characterize the performance of an algorithm in terms of the cumulative reward shortfall called regret.
A popular algorithm based on an upper confidence bound achieves $O(\log T)$ regret for bounded rewards using a concentration inequality~\citep{Auer2002}; this rate is optimal~\citep{lairobbins1985}.
Yet its extension to heavy-tailed rewards was not well known.
%Yet network measurements had reported long-tailed delays~\citep{papagiannaki2003measurement}, and the corpus did not establish the same guarantee assuming only uniformly bounded $(1+\epsilon)$-moments.
Bubeck et al.\ later filled this gap using robust estimators, and Agrawal et al.\ subsequently improved and clarified the connection to bounded-reward models~\citep{bubeck2012banditsheavytail,DBLP:conf/colt/AgrawalJK21}.
The research problem of characterizing regret under heavy-tailed rewards is what we call a \emph{structural gap}: a pattern that is present in the literature but not yet resolved.
%Thus the cutoff exposes a structural gap later resolved by human research.
SGHA emulates this process prospectively by detecting and verifying analogous corpus-level gaps and formulating them as evidence-grounded research problems.

Existing systems either condition hypothesis generation on a research background comprising an explicit research question and a survey of prior work~\citep{yang2026moosestar}, or formulate candidate research problems, directions, or hypotheses within a search space anchored by a human-authored code template (AI Scientist v1)~\citep{lu2024ai}, a broad topical description (AI Scientist v2)~\citep{yamada2025ai}, a scientist-specified research goal~\citep{gottweis2025towards}, a selected core paper~\citep{baek2025researchagent}, or an input topic used for graph-grounded literature retrieval~\citep{li2026graph2idea}. ResearchAgent explicitly includes an LLM-based problem-identification step, but generates the problem from a core paper, its citation neighborhood, and retrieved entities. Very recently, Graph2Idea retrieves papers according to an input topic, transforms them into structured knowledge triples, dynamically constructs a target-centered knowledge graph, and then plans research directions before synthesizing complete ideas~\citep{li2026graph2idea}. To our knowledge, these systems do not integrate all three capabilities: typed structural-gap detection over a topic-bounded full-text corpus, corpus-level verification before formulation, and passage-level provenance for evidence, counterevidence, and explicit ambiguities.
Beyond base-model capability, deployment is an independent practical consideration. In their reported implementations or principal experimental configurations, several prominent systems use proprietary models or hosted inference services for central idea generation or refinement~\citep{lu2024ai,baek2025researchagent,yamada2025ai,gottweis2025towards}, limiting exact reproducibility and model-level auditability relative to locally served open-weight inference. When unpublished or sensitive research materials are transmitted to external services, this additionally introduces deployment-dependent confidentiality, privacy, and data-governance concerns.

We introduce \emph{SGHA}, the Structural Gap Hypothesis Agent, a training-free system for generating candidate research-problem families from scientific papers. Figure~\ref{fig:sgha-overview} summarizes the pipeline. Given a topic, SGHA retrieves and parses relevant papers, extracts evidence-backed scientific tuples and paper-level objects, and builds a typed evidence graph linked to supporting passages. It then searches this graph for candidate structural gaps. Only gaps that pass a hard verification gate proceed to formulation. SGHA first writes a direct formulation close to the evidence, then generates broader variants and uses a separate critic to reject variants that are unsupported or only rhetorically stronger. Surviving formulations are grouped into project families and converted into semi-formal problem statements that describe the main variables, assumptions, feedback or observation model, objective, success criteria, and unresolved ambiguities. When the evidence is too weak, SGHA reports a low-signal outcome rather than forcing a research menu.

SGHA also supports two complementary modes. A profile-conditioned mode uses a researcher's literature context and an optional target topic to prioritize problems relevant to that context. An evolutionary exploration branch iteratively broadens literature-grounded seeds to produce additional candidate directions. These modes use the same underlying goal: to keep generated problems tied to evidence rather than treating them as unconstrained prompt completions.

We evaluate SGHA across five machine-learning domains: bandits, in-context learning, reasoning and test-time computation, offline reinforcement learning, and uncertainty estimation. The main experiments use a frozen local \texttt{Qwen/Qwen3.5-9B} model for generation, with no task-specific training or fine-tuning. Across 1,250 selected papers, SGHA extracts structured content from 1,044 papers. After verification, 39 candidate gaps remain and lead to 15 formalized project families. Our primary comparison is with the AI-Scientist-v2 ideation workflow configured with the same local model and matched output counts. This is a system-level comparison rather than an information-matched ablation: SGHA uses the bounded corpus, whereas the AI-Scientist-v2 retains its Semantic Scholar retrieval. Since SGHA is meant to write research-problem formulations rather than full project plans, we evaluate the comparison with a formulation-only rubric. Under this rubric, a method-label-masked evaluation using five frontier-model judges from different providers assigns SGHA higher mean scores on all but one of the formulation-quality criteria, including well-posedness, assumption-boundary clarity, formalizability, source-grounded specificity, and ambiguity handling. The local AI-Scientist-v2 baseline receives a higher mean score only on scope control, consistent with its tendency to produce narrower proposal-style ideas. The hosted judges are used only for post hoc evaluation after the candidate outputs have been frozen and are not part of SGHA's generation or verification pipeline. Additional comparisons using Claude Opus in the AI-Scientist-v2 workflow and the public MOOSE-Star HC model \cite{yang2026moosestar} were conducted.
In the Claude Opus diagnostic, which matches retained output counts but not compute or information budgets, SGHA receives a competitive numerical mean score on overall formulation quality to AI-Scientist-v2 with Claude Opus.

We additionally report exploratory diagnostics on profile conditioning, model configuration, corpus size, and evolutionary-output availability, treating them as capability or sensitivity analyses rather than controlled comparisons.

\textbf{Our main contributions are:}
\begin{contributionbox}
\begin{itemize}
    \item \textbf{Corpus-first research-problem discovery.}
    We present SGHA, an end-to-end pipeline that detects unresolved cross-paper patterns in a bounded corpus and converts verified gaps into evidence-grounded families of research problems. All generation and verification stages run on a locally served open-weight LLM without requiring proprietary frontier-model APIs.

    \item \textbf{Verification before formulation.}
    SGHA does not formulate a research problem from every detected pattern. It first screens each candidate for evidential support and prior coverage within the bounded corpus, formulates only those that pass verification, and uses a separate critic to reject unsupported ambition-expanded variants.

    \item \textbf{Traceable problem formulations.}
    SGHA outputs problem specifications with explicit assumptions, objectives, and success criteria. Each specification retains its source motif type and includes passage-level evidence and counterevidence links together with explicit ambiguity annotations.

    \item \textbf{Empirical analysis of quality, breadth, and failure modes.}
    We evaluate SGHA across multiple domains with a local model, compare it primarily with a locally configured AI-Scientist-v2 idea generation workflow, and report exploratory diagnostics on profile conditioning, cross-model behavior, corpus size, low-signal behavior, and evolutionary-exploration availability.
\end{itemize}
\end{contributionbox}

\begin{relatedworkbox}{Related Work on Automated Research Ideation with Local LLMs}
Open-weight local LLMs have been used in automated scientific ideation. SciMON fine-tunes T5-large and, in a biochemical case study, Meditron-7B on paper-derived supervision, while its reported iterative novelty-boosting experiments use GPT-4~\citep{wang-etal-2024-scimon}. HypER uses a Llama-3.1-70B teacher to construct and score citation chains and LoRA-tunes Phi-3 Mini 3.8B for relevance and chain-validation tasks; hypothesis generation is then performed at inference time, conditioned on validated chains~\citep{vasu-etal-2025-hyper}. MOOSE-Star fine-tunes R1-Distill-Qwen-7B models for inspiration retrieval and hypothesis composition using supervision generated with a locally deployed 32B teacher~\citep{yang2026moosestar}.
These locally executable open-weight instantiations rely on task-specific fine-tuning or distillation, and they do not explicitly model research-problem discovery as detecting and verifying unresolved structural gaps across a bounded literature corpus.
HypoGeniC can run with an open-weight Mixtral-8x7B model and updates a bank of predictive natural-language hypotheses from labeled examples without fine-tuning the generator; it does not search a literature corpus for research problems~\citep{zhou2024hypothesis}.

Among the most closely related prior works, Problem Discovery via Structural Motifs in Knowledge Graphs (PRISM)~\citep{gharat2026prism} explores structural-motif-based problem discovery by detecting structural gaps in a paper corpus, converting graph motifs into structured problem objects, and refining them with a novelty-and-feasibility critic. 
Its expert-evaluation protocol is proposed rather than carried out, leaving open whether detected motifs represent genuine unresolved gaps rather than retrieval or extraction artifacts or questions already covered elsewhere in the corpus. SGHA partially addresses this issue within the bounded corpus by verifying candidate gaps before formulation, merging redundant directions, and making remaining ambiguities explicit.
\end{relatedworkbox}

%% file: files/sgha.tex
\section{Structural Gap Hypothesis Agent}
\label{sec:sgha}

SGHA is designed around a simple constraint: a generated research problem should be tied to evidence in the literature. Given a topic, the system first builds a structured evidence base from papers, then looks for recurring patterns in that evidence, verifies the resulting candidate gaps, and only then formulates research-problem families. This is different from asking a language model to directly generate ideas from a topic. The language model is used in specific stages, while graph construction, motif detection, verification gating, family consolidation, and final auditing keep the pipeline grounded.

The overview figure in the introduction shows the full pipeline. In this section, we describe the main objects that move through it: evidence tuples, typed evidence graphs, structural motifs, candidate gaps, verified gaps, project families, and formal problem statements. We use one Bandits example as a running reference. In the selected corpus, SGHA extracts evidence that diffusion-based Thompson sampling (dTS) is analyzed under linear reward or link-function assumptions and bounded contexts, while related evidence points to nonlinear score functions requiring approximation. This trace later leads to the project family \emph{Characterizing the Non-Convex Failure Regime of Diffusion-Based Contextual Bandits}. Section~\ref{sec:qualitative-results} shows the final generated formulation, proposal-style abstract, formal problem object, risks, and ambiguity flags for this example.

\subsection{Building the Evidence Base}

SGHA begins by constructing a topic-specific corpus. We distinguish three paper-level stages. In stage one, \emph{selected} papers are chosen for the corpus before parsing. Next, \emph{parsed} papers are selected papers for which usable text is obtained, and finally, the \emph{extracted} papers are parsed papers for which SGHA produces a valid structured extraction.

For each extracted paper, SGHA records both paper-level objects and relation-level evidence. The paper-level object contains methods, tasks, datasets, metrics, assumptions, results, limitations, failure conditions, claims, contradictions or tensions, and future work. These fields are useful because not every important statement in a paper becomes a relation edge in the graph.

The relation-level representation consists of evidence tuples of the form
\[
(\text{subject}, \text{relation}, \text{object}),
\]
with source provenance. Relations include \texttt{assumes}, \texttt{fails\_under}, \texttt{limited\_by}, \texttt{addresses}, \texttt{not\_addressed\_by}, \texttt{contradicts}, \texttt{improves\_over}, \texttt{evaluated\_on}, \texttt{uses\_dataset}, and \texttt{measured\_by}. Each tuple also stores supporting text, paper section, claim type, polarity, condition, evidence type, confidence, and source-paper identifier.

These tuples are local evidence records extracted from individual papers. Before SGHA uses them downstream, it validates them by removing invalid relations, unsupported spans, missing evidence, generic subjects, and relation-specific errors. This keeps the evidence base from being dominated by vague labels such as ``our method'' or claims that are not supported by the parsed text.

These tuples are local evidence records extracted from individual papers. Before SGHA uses them in the graph, it applies deterministic validation and cleanup. First, the relation must be one of the allowed relation types, and the subject, object, and evidence fields must be present. Second, the supporting evidence span must match the parsed paper text after normalization, so that the tuple can be traced back to the source paper. Third, SGHA removes generic subjects such as "method", "algorithm", or "our method", because these labels are too vague to support graph construction. It also filters overly generic task labels, drops evaluation tuples that do not include task, dataset, or metric context, and normalizes metadata such as relation polarity. For example, \texttt{fails\_under} and \texttt{limited\_by} are treated as negative relations, while \texttt{assumes} is treated as neutral. These checks keep the evidence base tied to specific paper text and prevent vague or unsupported tuples from driving later gap detection.

\subsection{From Evidence Tuples to a Typed Graph}

SGHA then inserts the validated paper objects and evidence tuples into a typed evidence multigraph. The graph has different kinds of nodes for different scientific objects. Paper nodes represent source papers, while other nodes represent methods, assumptions, limitations, failure conditions, claims, tasks, datasets, metrics, and results. The edges come from the extracted tuples. For example, an edge may state that a method \texttt{assumes} an assumption, \texttt{fails\_under} a condition, or is \texttt{measured\_by} a metric. Each edge keeps its source provenance, including the paper identifier and supporting evidence span. Thus, the graph is not just a citation graph or a collection of paper summaries. It records the scientific relations that SGHA extracted from the corpus: what methods assume, where they are evaluated, what they improve over, where they fail, and which limitations remain unresolved.

\begin{figure*}[t]
  \centering
  \includegraphics[width=\textwidth]{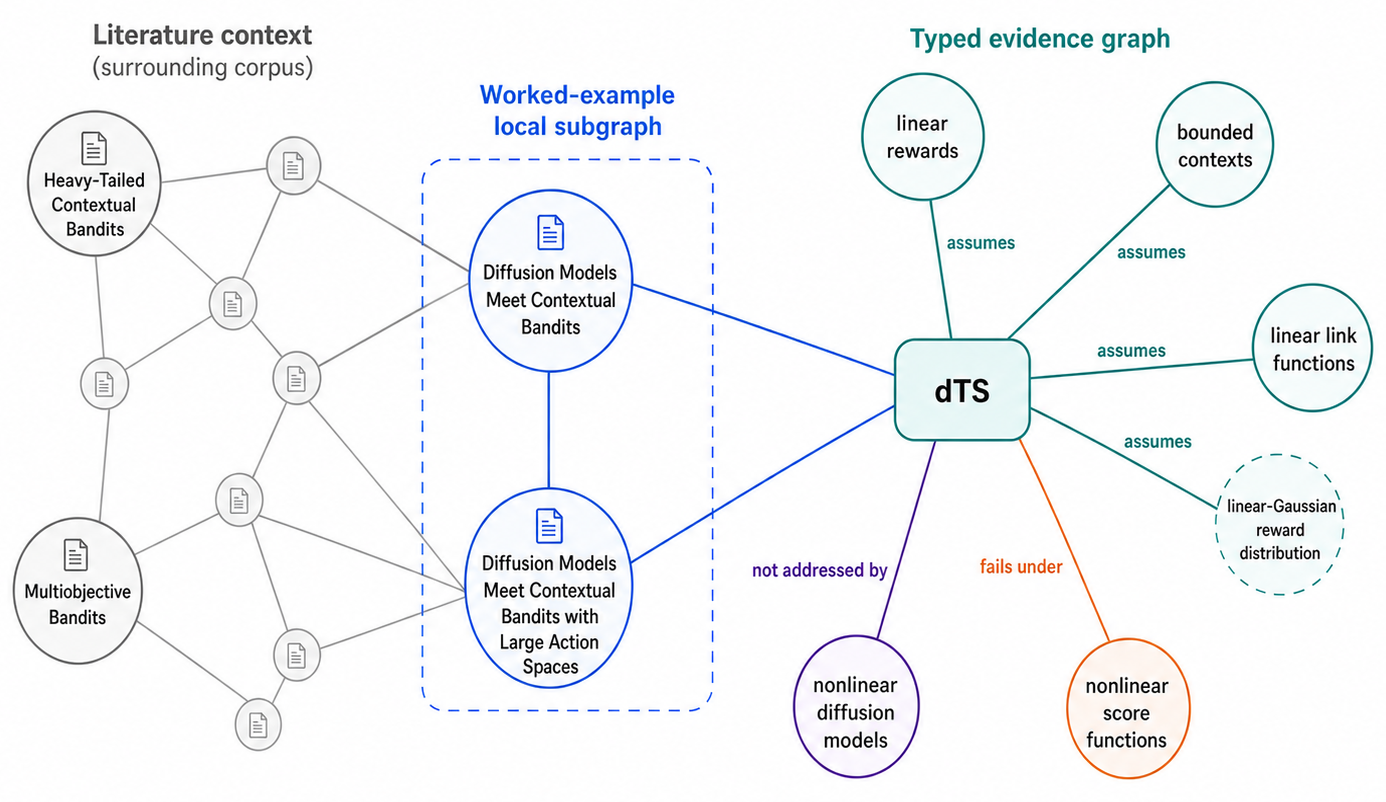}
  \caption{Evidence graph view of the Bandits worked example. The left side shows the surrounding literature context and the two source papers used in the worked example. The right side shows the typed local evidence graph: dTS is linked to linearity and bounded-context assumptions, an open issue around nonlinear diffusion models, and a nonlinear score-function failure condition. This local graph supports the assumption--failure candidate gap used later in the qualitative result.}
  \label{fig:evidence-graph-bandits}
\end{figure*}

Figure~\ref{fig:evidence-graph-bandits} shows this construction for the running Bandits example. The left side shows the surrounding literature context and highlights the two source papers \citep{aouali2024diffusion,aouali2026diffusion} used in the trace. The right side shows the typed local graph built from the extracted tuples. The method node dTS is connected to assumption nodes for linear rewards, bounded contexts, and linear link functions. It is also connected to an open issue around nonlinear diffusion models and to a failure-condition node for nonlinear score functions. This local graph is the evidence structure that later triggers the assumption--failure motif. 

\subsection{Structural Motifs and Candidate Gaps}

Once the evidence graph is built, SGHA searches it for predefined graph patterns, which we call structural motifs. A motif is a small pattern of relations that may point to a possible research gap. For example, a method may depend on an assumption in one part of the literature and appear to fail when a related condition changes in another part of the literature. When SGHA finds such a pattern, it creates a candidate structural gap. This candidate is not yet a verified research problem, and is only a place in the literature graph that deserves closer inspection.

\begin{figure*}[t]
  \centering
  \includegraphics[width=\textwidth]{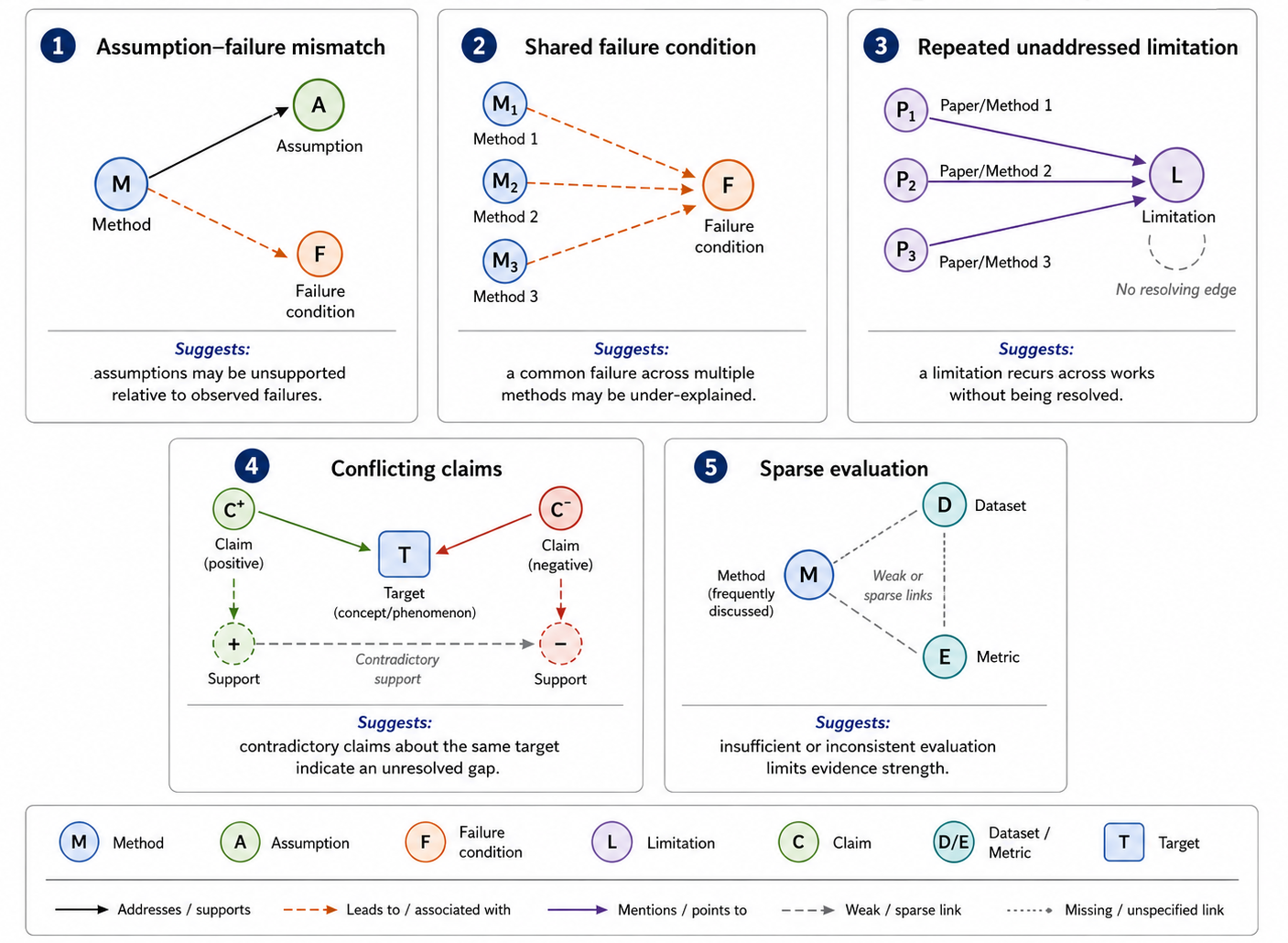}
  \caption{Examples of structural motifs used for candidate-gap discovery. Each mini-panel shows a deterministic graph pattern queried over the typed evidence graph. A motif creates a candidate gap, not a final research problem. SGHA later screens, verifies, formulates, and consolidates these candidates before they can enter the final output.}
  \label{fig:structural-motifs}
\end{figure*}

Figure~\ref{fig:structural-motifs} shows several motif families used by SGHA. An assumption--failure motif links a method to an assumption and to a related failure condition. A shared-failure motif appears when several methods fail under the same condition. A repeated-unaddressed-limitation motif appears when a limitation is mentioned across papers without evidence that it has been addressed. Other motifs capture conflicting claims, sparse evaluation, unrealistic assumptions, missing stress tests, and theory--practice gaps. These motifs are deliberately simple: their role is to find candidates for later screening and verification, not to decide on their own that a research problem is valid.

In the Bandits example, the local graph links dTS to assumptions that make posterior approximation tractable, and also to a nonlinear score-function condition where approximation becomes necessary. This matches the assumption--failure motif. The resulting candidate gap can thus be read as follows: dTS is analyzed under linear and bounded assumptions, but nearby nonlinear score-function settings may violate those assumptions. At this stage, SGHA has only identified a structured pattern in the literature. The candidate still needs to pass novelty screening and verification before it can be formulated as a research problem.

\subsection{Verification Before Formulation}

Such candidate gaps first enter a checking stage. The goal of this stage is to turn a graph-derived pattern into a formulation-ready gap. SGHA does this in two steps: a corpus-level screen followed by role-based verification.

The corpus-level screen keeps candidates that remain meaningful after checking nearby evidence in the selected corpus. Candidates retained by this screen are called \emph{novelty survivors}. In SGHA, this term has a specific meaning: the candidate has passed the system's internal corpus screen and is worth deeper review.

SGHA then sends a selected subset of novelty survivors to role-based verification. These are the \emph{reviewed gaps}. The verification agents examine each reviewed gap from complementary perspectives. The support agent identifies evidence that the gap is grounded in the corpus. The skeptic agent looks for nearby counterevidence or possible resolutions. The feasibility agent checks whether the problem can plausibly be studied. The mechanism agent asks whether there is a reasonable explanation for the gap. The verification critic gives a conservative final assessment.

A reviewed gap becomes a \emph{verified gap} when it passes the hard verification gate. This gate is deterministic. It admits candidates with the required agent outputs, no disqualifying parse failures, a survival score above threshold, and a non-rejecting critic with sufficient confidence. Verified gaps, then enter the direct-formulation stage.

In the running Bandits example, this stage keeps the assumption--failure pattern grounded but also sharpens its interpretation. The support agent finds evidence for the linear-assumption issue in dTS. The skeptic and critic add useful context: approximation-based approaches are mentioned in the source papers, so the final problem should frame the issue carefully rather than overstate it. The feasibility and mechanism agents find the problem studyable and mechanistically plausible. The hard gate passes the candidate, allowing SGHA to formulate it. In this way, verification turns the graph-derived dTS candidate into a formulation-ready gap.

\subsection{From Verified Gaps to Project Families}

Once a gap is verified, SGHA treats it as an evidence-backed starting point for formulation. The first step is a direct formulation. This formulation stays close to the verified gap: it states the main setting, the central assumption or failure, the objective, and a plausible theorem, algorithm, or benchmark target. It also keeps the source gap, motif type, supporting papers, and verification provenance attached to the formulation.

In the running Bandits example, the direct formulation is \emph{Extending dTS to Non-Linear Score Functions in Contextual Bandits}. This is intentionally close to the verified evidence. It asks how diffusion-based posterior inference can be extended beyond linear score functions while preserving useful regret guarantees or approximation behavior.

SGHA then broadens the direct formulation in a controlled way. It generates conservative, generalized, and bold variants. The conservative variant makes a small extension of the direct problem. The generalized variant broadens the problem class or assumptions while keeping the formulation grounded. The bold variant asks whether the evidence points to a deeper boundary, impossibility, or failure-regime question.

This expansion stage is paired with an independent formulation critic. The critic checks whether a variant makes a real scientific move, rather than only sounding more ambitious. It looks for source support, specificity, non-incrementality, and a clear change in the problem class or objective. Variants that pass this critic become candidates for final project families.

In the Bandits example, this step changes the formulation from a local extension of dTS to a broader boundary question: \emph{Characterizing the Non-Convex Failure Regime of Diffusion-Based Contextual Bandits}. The accepted formulation no longer asks only how to patch dTS for nonlinear score functions. It asks when diffusion-based inference itself fails under nonlinear score landscapes.

Finally, SGHA consolidates critic-passing variants into project families. This step avoids presenting several versions of the same underlying direction as separate outputs. Variants can be grouped when they share the same verified gap, direct formulation, supporting papers, or research question. Each family keeps a representative formulation, member variants, source verified gaps, source direct formulations, supporting papers, critic rationale, risks, and internal quality metadata.

\subsection{Formal Problem Statements}

After project families are formed, SGHA turns each family into a semi-formal problem object. This final step makes the research direction easier to inspect. The problem object records the main entities, variables, observations or feedback, decision outputs, objective, constraints, success criterion, assumptions, possible result types, source grounding, and ambiguity flags. The aim is to give a researcher a clear starting point: what is already specified, what the candidate is trying to study, and what still needs to be defined more carefully.

In the Bandits example, the final family is formalized using a score-function class \(\mathcal{F}\), a diffusion-based inference algorithm \(\pi_{\mathrm{diff}}\), true and approximate reward distributions, a score landscape, and cumulative regret \(\mathcal{R}_T\). The resulting problem asks when diffusion-based inference can guarantee sublinear regret, and where this guarantee breaks down because the Gaussian approximation diverges from the true reward distribution under nonlinear score landscapes.

SGHA also records the parts of the formulation that remain underspecified. For this example, it flags the meaning of non-convexity in the score landscape, the operational definition of identification failure, and the precise threshold at which approximation error becomes a failure. These ambiguity flags are part of the final artifact. They make clear which definitions a researcher would need to sharpen before turning the candidate into a theorem, algorithm, or benchmark study.

\subsection{Profile Conditioning, Evolutionary Exploration, and Low-Signal Outcomes}

The same SGHA pipeline can also be used in a profile-conditioned setting. Here, the input includes a researcher's literature context and, optionally, a target topic. This information is used during corpus construction and prioritization, so the evidence base is closer to the researcher's previous work or stated interests. The downstream checks remain the same: candidate gaps are still detected from the evidence graph, screened, verified, formulated, critiqued, consolidated, and formalized. Thus, profile conditioning changes what evidence the system starts from, but not the standards a candidate must satisfy before becoming a final project family.

\begin{figure*}[t]
  \centering
  \includegraphics[width=\textwidth]{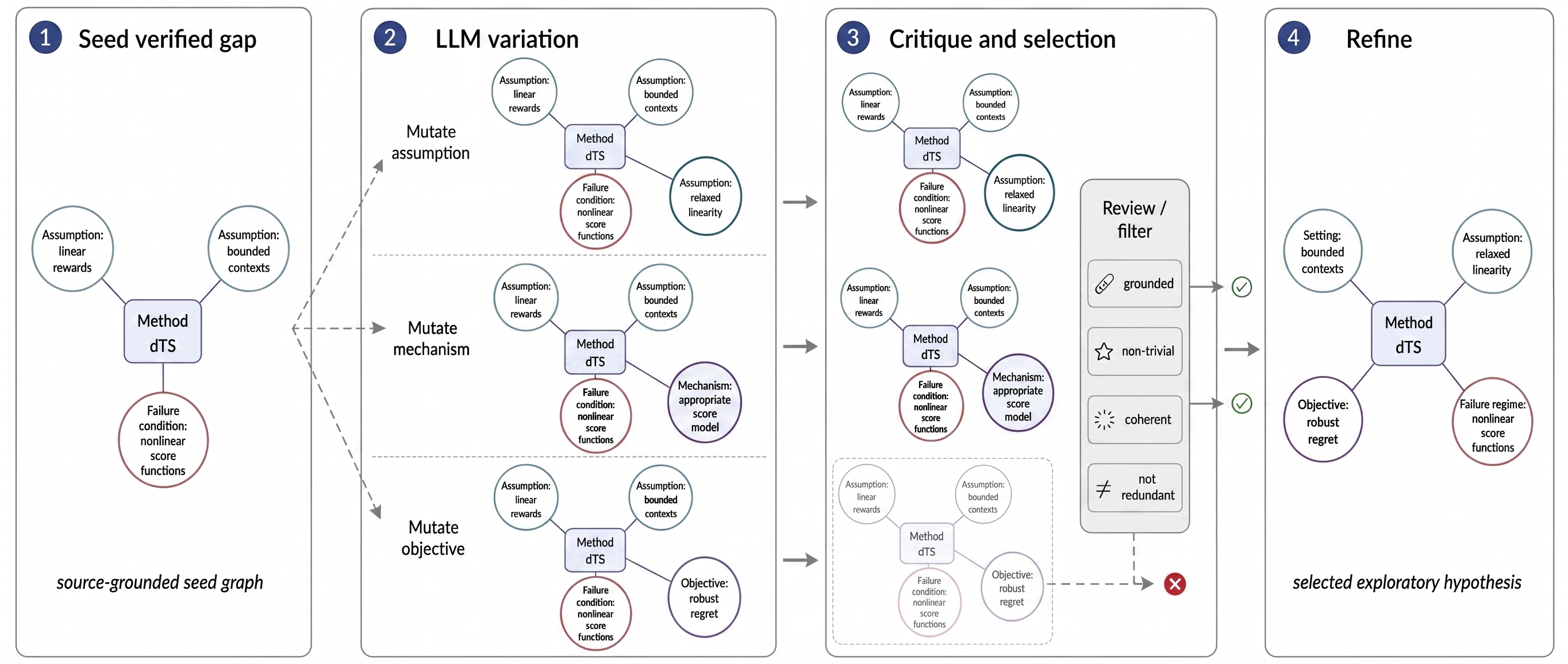}
  \caption{Graph-mutation view of the optional evolutionary exploration branch. Starting from a literature-grounded seed gap, the branch proposes variants by changing assumptions, mechanisms, objectives, or regimes. Variants are critiqued, selected, and refined before a new exploratory hypothesis is retained. This branch is used for breadth and is kept separate from the main verification-gated family-report path unless stated otherwise.}
  \label{fig:evolutionary-graph-mutation}
\end{figure*}

SGHA also includes an optional evolutionary exploration branch. This branch is separate from the main verification-gated family-report path. It starts from a literature-grounded seed, such as a verified gap or an early formulation, and explores nearby problem variants by changing assumptions, mechanisms, objectives, or regimes. The variants are then critiqued, selected, and refined over several rounds. Figure~\ref{fig:evolutionary-graph-mutation} illustrates this view: evolutionary exploration can be seen as mutating a graph-level problem representation while preserving its connection to the original evidence.

The purpose of this branch is breadth. The main SGHA path is conservative: it produces a compact set of verified, critic-filtered, and formalized project families. The evolutionary branch instead explores a wider neighborhood of possible directions around the same evidence base. For this reason, unless explicitly stated, evolutionary outputs are reported separately from the main project-family counts and from the main formulation-quality comparison.

Finally, SGHA can report a low-signal outcome when the evidence does not support a final research menu. This can happen when no candidate passes verification, when no ambition-expanded variant passes the critic, or when no project family survives consolidation. In such cases, the system completes the run but reports that the corpus did not yield a sufficiently supported final set of project families. For a literature-driven system, this behavior is important: SGHA should surface grounded opportunities when the evidence supports them, and avoid forcing unsupported directions when it does not.

%% file: files/experimental_setup.tex
\section{Experimental Setup}
\label{sec:experimental-setup}

We evaluate SGHA as a system for producing research-problem formulations from scientific papers. The experiments are organized around three questions. First, can SGHA produce final project-family artifacts across different areas of machine learning? Second, how do these artifacts compare with outputs from automated ideation baselines under a common formulation-quality rubric? Third, how does the system behave in additional settings, including profile-conditioned generation, evolutionary exploration, and Bandits sensitivity analyses over model size and corpus size?

The main SGHA experiments use a frozen local \texttt{Qwen/Qwen3.5-9B} model, with no task-specific training or fine-tuning. We compare SGHA with output-count-matched baselines and evaluate all retained candidates with the same blinded formulation-only judge. Alongside judge scores, we also report structural properties of the generated artifacts, such as source grounding, assumptions, formal problem statements, risks, and ambiguity flags.

\subsection{Domains and Corpora}

We run SGHA on five machine-learning domains: bandits, in-context learning, reasoning and test-time computation, offline reinforcement learning, and uncertainty estimation. These domains cover different parts of ML research, from sequential decision-making and in-context behavior to test-time reasoning, offline policy learning, and uncertainty quantification. Each domain uses a budget of 250 selected papers. To include both topical and source diversity, the bandits, in-context learning, and reasoning/test-time-computation corpora are drawn from OpenReview, while the offline-reinforcement-learning and uncertainty-estimation corpora are drawn from arXiv.

Together, these five corpora contain 1,250 selected papers. SGHA parses 1,049 of them, meaning that usable text is obtained, and produces valid structured scientific extractions for 1,044 papers. These extractions yield 8,634 evidence tuples, which are used to build typed evidence graphs and instantiate candidate structural gaps. Table~\ref{tab:experimental-corpus-counts} reports the corpus and extraction counts by domain.

% \begin{table*}[t]
% \centering
% \small
% \caption{Corpus and extraction counts for the five main SGHA runs. ``Parsed'' denotes papers for which usable text is obtained; ``Extracted'' denotes papers for which SGHA produces a valid structured extraction.}
% \label{tab:experimental-corpus-counts}
% \begin{tabular}{p{0.30\linewidth}lrrrr}
% \toprule
% Domain & Source & Selected & Parsed & Extracted & Tuples \\
% \midrule
% Bandits & OpenReview & 250 & 221 & 221 & 1554 \\
% In-context learning & OpenReview & 250 & 193 & 193 & 1559 \\
% Reasoning / test-time computation & OpenReview & 250 & 137 & 137 & 1139 \\
% Offline reinforcement learning & arXiv & 250 & 248 & 245 & 2176 \\
% Uncertainty estimation & arXiv & 250 & 250 & 248 & 2206 \\
% \midrule
% Total & -- & 1250 & 1049 & 1044 & 8634 \\
% \bottomrule
% \end{tabular}
% \end{table*}

\begin{table*}[t]
\centering
\small
\caption{Corpus and extraction counts for the five main SGHA runs. ``Parsed'' denotes papers for which usable text is obtained; ``Extracted'' denotes papers for which SGHA produces a valid structured extraction.}
\label{tab:experimental-corpus-counts}
\begingroup
\renewcommand{\arraystretch}{1.12}
\begin{tabular}{L{0.30\linewidth}L{0.15\linewidth}rrrr}
\toprule
\tableheader
\theadcell{Domain} & \theadcell{Source} & \theadcell{Selected} & \theadcell{Parsed} & \theadcell{Extracted} & \theadcell{Tuples} \\
\midrule
\textbf{Bandits} & OpenReview & 250 & 221 & 221 & 1554 \\
\textbf{In-context learning} & OpenReview & 250 & 193 & 193 & 1559 \\
\textbf{Reasoning / test-time computation} & OpenReview & 250 & 137 & 137 & 1139 \\
\textbf{Offline reinforcement learning} & arXiv & 250 & 248 & 245 & 2176 \\
\textbf{Uncertainty estimation} & arXiv & 250 & 250 & 248 & 2206 \\
\midrule
\tabtotal
\textbf{Total} & -- & \textbf{1250} & \textbf{1049} & \textbf{1044} & \textbf{8634} \\
\bottomrule
\end{tabular}
\endgroup
\end{table*}

\subsection{Systems Compared}

The SGHA outputs in the main comparison come from the verification-gated family-report path described in Section~\ref{sec:sgha}. The candidate gaps are detected from the evidence graph, screened against nearby corpus evidence, passed through the hard verification gate, formulated, expanded, filtered by an independent critic, consolidated into project families, and converted into semi-formal problem statements. The evaluated SGHA candidate is the final project-family artifact, rather than an intermediate motif hit, direct formulation, or rejected variant.

Our primary baseline is native AI-Scientist-v2 ideation with the same local \texttt{Qwen/Qwen3.5-9B} model. We use its ideation entry point and match the number of retained candidates to the number of SGHA final families in each domain. This gives 15 candidates from each method: 3 for bandits, 4 for in-context learning, 1 for reasoning and test-time computation, 1 for offline reinforcement learning, and 6 for uncertainty estimation.

We also include two reference baselines. The first is AI-Scientist-v2 with \texttt{anthropic/claude-opus-latest} as a stronger generator, again run in ideation-only mode and matched to SGHA's retained candidate counts. The second is MOOSE-Star using the released \texttt{ZonglinY/MOOSE-Star-HC-R1D-7B} public model in HC-only mode, with no additional training or fine-tuning. These baselines give reference points for same-model native ideation, stronger-generator ideation, and public trained hypothesis generation.

% \begin{table*}[t]
% \centering
% \small
% \caption{Systems compared in the formulation-quality evaluation. All methods are output-count matched to 15 retained candidates across the five domains.}
% \label{tab:experimental-methods}
% \begin{tabular}{p{0.24\linewidth}p{0.23\linewidth}p{0.20\linewidth}p{0.23\linewidth}}
% \toprule
% Method & Model & Mode & Evaluated artifact \\
% \midrule
% SGHA &
% \texttt{Qwen3.5-9B} &
% Verification-gated literature-gap pipeline &
% Final project-family artifact with provenance and a semi-formal problem statement. \\

% AI-Scientist-v2 + Qwen &
% \texttt{Qwen3.5-9B} &
% Native ideation-only workflow &
% Retained generated idea normalized into the candidate-packet format. \\

% AI-Scientist-v2 + Claude Opus &
% \texttt{claude-opus-latest} &
% Native ideation-only workflow &
% Retained generated idea from a stronger generator setting. \\

% MOOSE-Star &
% \texttt{MOOSE-Star-HC-R1D-7B} &
% Released public HC-only model &
% Retained generated hypothesis from the public model. \\
% \bottomrule
% \end{tabular}
% \end{table*}

\begin{table*}[t]
\centering
\small
\caption{Systems compared in the formulation-quality evaluation. All methods are output-count matched to 15 retained candidates across the five domains.}
\label{tab:experimental-methods}
\begingroup
\renewcommand{\arraystretch}{1.16}
\begin{tabular}{L{0.23\linewidth}L{0.20\linewidth}L{0.22\linewidth}L{0.27\linewidth}}
\toprule
\tableheader
\theadcell{Method} & \theadcell{Model} & \theadcell{Mode} & \theadcell{Evaluated artifact} \\
\midrule
\textbf{SGHA} &
\texttt{Qwen3.5-9B} &
Verification-gated literature-gap pipeline &
Final project-family artifact with provenance and a semi-formal problem statement. \\

\addlinespace[0.25em]
\textbf{AI-Scientist-v2 + Qwen} &
\texttt{Qwen3.5-9B} &
Native ideation-only workflow &
Retained generated idea normalized into the candidate-packet format. \\

\addlinespace[0.25em]
\textbf{AI-Scientist-v2 + Claude Opus} &
\texttt{claude-opus-latest} &
Native ideation-only workflow &
Retained generated idea from a stronger generator setting. \\

\addlinespace[0.25em]
\textbf{MOOSE-Star} &
\texttt{MOOSE-Star-HC-R1D-7B} &
Released public HC-only model &
Retained generated hypothesis from the public model. \\
\bottomrule
\end{tabular}
\endgroup
\end{table*}

Before judging, all outputs are converted into a common candidate format. The packet contains the title, problem statement, motivation or abstract, proposed direction, expected contribution, source or context information when available, assumptions or setup when available, formal-problem fields when available, ambiguity or missing-definition fields when available, and risks or caveats when available. Fields that a method does not produce are left as \texttt{not provided}; we evaluate the artifacts as produced and report structural-field coverage separately.

\subsection{Formulation-Only Evaluation}

SGHA produces research-problem formulations rather than complete project plans. We therefore evaluate candidates with a formulation-only rubric. The rubric focuses on whether a candidate states a clear problem, gives enough technical structure, controls its scope, makes assumptions visible, and identifies what remains ambiguous. Table~\ref{tab:formulation-rubric} lists the ten criteria used by the judge.

\begin{table*}[t]
\centering
\small
\caption{Formulation-only evaluation rubric. Each criterion is scored on a 0--10 scale.}
\label{tab:formulation-rubric}
\begingroup
\renewcommand{\arraystretch}{1.13}
\begin{tabular}{L{0.28\linewidth}L{0.62\linewidth}}
\toprule
\tableheader
\theadcell{Criterion} & \theadcell{What the judge evaluates} \\
\midrule
\textbf{Problem-definition clarity} & Whether the candidate states a clear research problem rather than a broad topic or loose motivation. \\
\textbf{Technical specificity} & Whether the formulation names concrete objects, settings, mechanisms, assumptions, metrics, or target results. \\
\textbf{Well-posedness} & Whether the problem has enough structure to be studied or refined. \\
\textbf{Assumption-boundary clarity} & Whether the formulation makes clear which assumptions are used, relaxed, questioned, or missing. \\
\textbf{Formalizability} & Whether the problem can plausibly be written as a theorem, algorithmic objective, benchmark protocol, impossibility result, or other formal research target. \\
\textbf{Nontriviality} & Whether the candidate goes beyond a simple application, small robustness check, or generic ``apply X to Y'' idea. \\
\textbf{Scope control} & Whether the problem is focused enough to be studied rather than combining many loosely related concepts. \\
\textbf{Source-grounded specificity} & Whether the formulation is tied to provided source context, evidence, or literature-derived details. \\
\textbf{Ambiguity hygiene} & Whether the candidate explicitly states missing definitions, unclear feedback models, unresolved assumptions, or caveats. \\
\textbf{Overall formulation quality} & The judge's overall assessment of the research-problem formulation under the rubric. \\
\bottomrule
\end{tabular}
\endgroup
\end{table*}

The 0--10 scores use the anchors in Table~\ref{tab:formulation-score-anchors}. These anchors are meant to make the scores interpretable: a score around 5 or 6 means the formulation is plausible but still needs refinement, while scores of 7 or 8 indicate a strong formulation.

% \begin{table}[t]
% \centering
% \small
% \caption{Score anchors used by the formulation-only judges.}
% \label{tab:formulation-score-anchors}
% \begin{tabular}{cl}
% \toprule
% Score & Interpretation \\
% \midrule
% 0 & No usable formulation. \\
% 1 & Mostly incoherent or unrelated to a research problem. \\
% 2 & Topic-level idea with almost no problem structure. \\
% 3 & Very vague formulation with major missing pieces. \\
% 4 & Weak formulation; some direction is visible, but the problem is poorly specified. \\
% 5 & Plausible idea, but important assumptions, scope, or definitions are missing. \\
% 6 & Plausible formulation with useful structure, but still needing substantial refinement. \\
% 7 & Strong formulation with clear problem structure and reasonable technical specificity. \\
% 8 & Very strong formulation; well posed, grounded, and mostly ready for expert refinement. \\
% 9 & Excellent formulation with unusually clear assumptions, scope, and formal target. \\
% 10 & Exceptional formulation; clear, grounded, formalizable, and close to research-ready. \\
% \bottomrule
% \end{tabular}
% \end{table}
\begin{table}[t]
\centering
\small
\caption{Score anchors used by the formulation-only judges.}
\label{tab:formulation-score-anchors}
\begingroup
\renewcommand{\arraystretch}{1.12}
\begin{tabular}{C{0.12\linewidth}L{0.74\linewidth}}
\toprule
\tableheader
\theadcell{Score} & \theadcell{Interpretation} \\
\midrule
\textbf{0} & No usable formulation. \\
\textbf{1} & Mostly incoherent or unrelated to a research problem. \\
\textbf{2} & Topic-level idea with almost no problem structure. \\
\textbf{3} & Very vague formulation with major missing pieces. \\
\textbf{4} & Weak formulation; some direction is visible, but the problem is poorly specified. \\
\textbf{5} & Plausible idea, but important assumptions, scope, or definitions are missing. \\
\textbf{6} & Plausible formulation with useful structure, but still needing substantial refinement. \\
\textbf{7} & Strong formulation with clear problem structure and reasonable technical specificity. \\
\textbf{8} & Very strong formulation; well posed, grounded, and mostly ready for expert refinement. \\
\textbf{9} & Excellent formulation with unusually clear assumptions, scope, and formal target. \\
\textbf{10} & Exceptional formulation; clear, grounded, formalizable, and close to research-ready. \\
\bottomrule
\end{tabular}
\endgroup
\end{table}

The evaluation uses method-label-masked candidate packets. Method labels are hidden from the judges, and the label key is used only during postprocessing. Judges evaluate the provided candidate text and supporting context under the fixed rubric; external literature novelty and correctness are outside the judge task.

To reduce dependence on a single model or provider, we use five LLM judges from different providers: \texttt{anthropic/claude-sonnet-4}, \texttt{openai/gpt-5.6-sol-pro}, \texttt{x-ai/grok-4.5}, \texttt{moonshotai/kimi-k3}, and \texttt{google/gemini-3.6-flash}. Each judge receives the same blinded packet format, rubric, score anchors, and response schema. The judge panel is used only after all candidate outputs have been frozen; it is not part of SGHA's generation or verification pipeline.

We report criterion-level means together with the judges' overall formulation-quality score. These scores are descriptive assessments of formulation quality, and we interpret them alongside structural properties of the generated outputs.
\subsection{Profile-Conditioned Generation}

We also evaluate a profile-conditioned version of SGHA. In this setting, the input includes a researcher-specific literature context and, optionally, a target topic. The profile context guides corpus construction and prioritization, so the selected papers and candidate gaps are closer to the researcher's previous work or stated area of interest.

After corpus construction, the downstream SGHA pipeline remains the same. Candidate gaps are detected from the evidence graph, screened, verified, formulated, expanded, filtered by the critic, consolidated into project families, and converted into semi-formal problem statements. Profile conditioning changes the evidence base and prioritization, while preserving the same checks before finalization.

We evaluate profile-conditioned generation on three researcher contexts: Yann LeCun, Geoffrey Hinton, and Michael I. Jordan. The resulting candidates are evaluated with a personalized judge rubric that includes the formulation-quality criteria from the main evaluation and additional profile-specific criteria for alignment, specificity, and fit to the provided profile context.

\subsection{Evolutionary Exploration}

SGHA includes an evolutionary exploration branch that serves a different role from the compact family-report path. The family-report path starts from verification-passed gaps and produces a small set of critic-filtered project families. The evolutionary branch explores a broader neighborhood around literature-grounded seeds.

Starting from a seed gap or formulation, the evolutionary branch varies parts of the problem representation, such as assumptions, mechanisms, objectives, or regimes. Candidate variants are critiqued, selected, and refined over multiple rounds. For evaluation, selected or ranked evolutionary hypotheses are normalized into the same candidate format used for the other methods and scored with the same formulation-only rubric. We report these outputs separately from the main SGHA project-family counts.

\subsection{Model- and Corpus-Size Sensitivity}

Finally, we include two sensitivity analyses on the Bandits domain. The model-size analysis compares the main \texttt{Qwen/Qwen3.5-9B} run with a smaller 4B configuration and a larger 27B configuration. This tests how extraction quality, verification yield, critic-passing variants, and final family construction change with model capacity.

The corpus-size analysis varies the number of selected Bandits papers. We run settings with 50, 100, 150, 200, and 250 selected papers. This tests how SGHA behaves as the evidence base grows. We treat this as a sensitivity study rather than a scaling-law experiment: larger corpora provide more evidence and more candidate gaps, while later stages such as verification, critic filtering, and family consolidation determine which candidates survive.

%% file: files/results.tex
\section{Results}
\label{sec:results}

\subsection{End-to-End Yield Across Domains}
\label{sec:main-pipeline-yield}

We first evaluate whether SGHA can complete the full path from papers to final research-problem families across different areas of machine learning. Table~\ref{tab:main-pipeline-yield} summarizes the five main runs. Across 1,250 selected papers, SGHA extracts structured scientific content from 1,044 papers and produces 8,634 evidence tuples. These evidence tuples support graph construction and candidate-gap discovery. After corpus screening and verification, 39 gaps are accepted for formulation. The final synthesis stages then produce 19 accepted variants, which are consolidated into 15 project families with 15 semi-formal problem statements.

This yield shows the intended behavior of the pipeline. SGHA starts from a large literature-derived evidence base and progressively refines it into a compact set of research-problem artifacts. The intermediate stages provide breadth: extraction creates evidence tuples, motifs surface candidate gaps, and verification identifies formulation-ready gaps. The final stages provide structure: ambition expansion, criticism, family consolidation, and formalization turn selected gaps into inspectable project families.

% \begin{table*}[t]
% \centering
% \small
% \caption{End-to-end SGHA yield across the five main domains. Verified gaps are candidate gaps that pass the hard verification gate. Accepted variants are ambition-expanded formulations that pass the independent critic.}
% \label{tab:main-pipeline-yield}
% \begin{tabular}{p{0.30\linewidth}rrrrr}
% \toprule
% Domain & Tuples & Verified gaps & Accepted variants & Families & Formal statements \\
% \midrule
% Bandits & 1554 & 6 & 4 & 3 & 3 \\
% In-context learning & 1559 & 12 & 4 & 4 & 4 \\
% Reasoning / test-time computation & 1139 & 7 & 1 & 1 & 1 \\
% Offline reinforcement learning & 2176 & 2 & 1 & 1 & 1 \\
% Uncertainty estimation & 2206 & 12 & 9 & 6 & 6 \\
% \midrule
% Total & 8634 & 39 & 19 & 15 & 15 \\
% \bottomrule
% \end{tabular}
% \end{table*}
\begin{table*}[t]
\centering
\small
\caption{End-to-end SGHA yield across the five main domains. Verified gaps are candidate gaps that pass the hard verification gate. Accepted variants are ambition-expanded formulations that pass the independent critic.}
\label{tab:main-pipeline-yield}
\begingroup
\renewcommand{\arraystretch}{1.12}
\begin{tabular}{L{0.30\linewidth}rrrrr}
\toprule
\tableheader
\theadcell{Domain} & \theadcell{Tuples} & \theadcell{Verified gaps} & \theadcell{Accepted variants} & \theadcell{Families} & \theadcell{Formal statements} \\
\midrule
Bandits & 1554 & 6 & 4 & 3 & 3 \\
In-context learning & 1559 & 12 & 4 & 4 & 4 \\
Reasoning / test-time computation & 1139 & 7 & 1 & 1 & 1 \\
Offline reinforcement learning & 2176 & 2 & 1 & 1 & 1 \\
Uncertainty estimation & 2206 & 12 & 9 & 6 & 6 \\
\midrule
\tabtotal
\textbf{Total} & \textbf{8634} & \textbf{39} & \textbf{19} & \textbf{15} & \textbf{15} \\
\bottomrule
\end{tabular}
\endgroup
\end{table*}

All five domains reach the final project-family stage, but with different yields. This is expected for a literature-driven system: some corpora contain several separable directions, while others lead to a smaller and tighter set of families. Uncertainty estimation produces the largest set of final families, while reasoning/test-time computation and offline reinforcement learning produce more focused outputs.

Table~\ref{tab:main-pipeline-yield} therefore defines the SGHA candidate set used in the main comparison: 15 formalized project families obtained from 39 verification-passed gaps. We next evaluate the quality of these artifacts---how clear, grounded, formalizable, and explicit about ambiguity they are---against automated ideation baselines.

\subsection{Formulation Quality Compared with Baselines}
\label{sec:formulation-quality-results}

In Table~\ref{tab:formulation-quality-results} we report the mean formulation-quality scores across the five LLM judges. SGHA obtains the highest overall formulation-quality score. The strongest gains are on the criteria that the pipeline is designed to improve: source-grounded specificity, ambiguity hygiene, assumption-boundary clarity, well-posedness, and formalizability.

The comparison with AI-Scientist-v2 using the same local Qwen model shows the clearest difference. Both systems use the same generator, but SGHA first builds an evidence graph, verifies candidate gaps, and produces formal problem objects. This leads to higher scores on nearly all formulation criteria, especially ambiguity hygiene and assumption-boundary clarity. The Claude Opus AI-Scientist-v2 baseline is much stronger on surface formulation: it produces clearer, more technically detailed, and more tightly scoped ideas. However, SGHA remains higher on the evidence-linked and structure-oriented criteria, and also has the highest overall score. MOOSE-Star scores lower in this setup, reflecting that the released HC-only model produces hypothesis text but not the same kind of verified, formalized problem artifact.

% \begin{table*}[t]
% \centering
% \scriptsize
% \caption{Formulation-quality scores averaged over five LLM judges. All methods are evaluated on 15 retained candidates. Bold indicates the best score in each column.}
% \label{tab:formulation-quality-results}
% \resizebox{\textwidth}{!}{
% \begin{tabular}{lrrrrrrrrrr}
% \toprule
% Method & Overall & Clarity & Technical & Well-posed & Boundary & Formal. & Nontrivial & Scope & Grounding & Ambiguity \\
% \midrule
% SGHA &
% \textbf{5.99} & 6.77 & 5.89 & \textbf{5.77} & \textbf{6.45} & \textbf{5.51} & 6.39 & 5.36 & \textbf{7.43} & \textbf{7.61} \\

% AI-Scientist-v2 + Qwen &
% 4.55 & 5.59 & 4.72 & 3.53 & 3.61 & 3.68 & 5.27 & 5.73 & 5.11 & 2.81 \\

% AI-Scientist-v2 + Claude Opus &
% 5.84 & \textbf{7.15} & \textbf{6.95} & 4.88 & 4.75 & 5.21 & \textbf{6.88} & \textbf{6.19} & 2.97 & 3.57 \\

% MOOSE-Star &
% 2.00 & 1.85 & 2.51 & 1.52 & 1.37 & 1.51 & 2.89 & 3.07 & 3.75 & 1.48 \\
% \bottomrule
% \end{tabular}
% }
% \end{table*}

\begin{table*}[t]
\centering
\scriptsize
\caption{Formulation-quality scores averaged over five LLM judges. All methods are evaluated on 15 retained candidates. Bold indicates the best score in each column.}
\label{tab:formulation-quality-results}
\begingroup
\renewcommand{\arraystretch}{1.14}
\resizebox{\textwidth}{!}{
\begin{tabular}{L{0.23\linewidth}rrrrrrrrrr}
\toprule
\tableheader
\theadcell{Method} & \theadcell{Overall} & \theadcell{Clarity} & \theadcell{Technical} & \theadcell{Well-posed} & \theadcell{Boundary} & \theadcell{Formal.} & \theadcell{Nontrivial} & \theadcell{Scope} & \theadcell{Grounding} & \theadcell{Ambiguity} \\
\midrule
\textbf{SGHA} &
\bestscore{5.99} & 6.77 & 5.89 & \bestscore{5.77} & \bestscore{6.45} & \bestscore{5.51} & 6.39 & 5.36 & \bestscore{7.43} & \bestscore{7.61} \\

AI-Scientist-v2 + Qwen &
4.55 & 5.59 & 4.72 & 3.53 & 3.61 & 3.68 & 5.27 & 5.73 & 5.11 & 2.81 \\

AI-Scientist-v2 + Claude Opus &
5.84 & \bestscore{7.15} & \bestscore{6.95} & 4.88 & 4.75 & 5.21 & \bestscore{6.88} & \bestscore{6.19} & 2.97 & 3.57 \\

MOOSE-Star &
2.00 & 1.85 & 2.51 & 1.52 & 1.37 & 1.51 & 2.89 & 3.07 & 3.75 & 1.48 \\
\bottomrule
\end{tabular}
}
\endgroup
\end{table*}

The table shows a useful tradeoff. Strong ideation models can produce polished and well-scoped ideas, as reflected by the Claude Opus baseline. SGHA is strongest when the rubric asks for problem structure that depends on the pipeline: grounding in the provided literature, explicit assumptions, formalizability, and clean handling of ambiguity. This supports the central claim of the paper: the benefit of SGHA is not only that it generates plausible ideas, but that it turns literature-derived gaps into inspectable research-problem formulations.

\subsection{Structural Artifact Coverage}
\label{sec:structural-coverage}

The judge scores measure formulation quality, but they do not fully show what information is available in each generated artifact. We therefore also inspect the structure of the outputs directly. Table~\ref{tab:structural-artifact-coverage} reports whether each candidate includes the fields that make a research-problem artifact easier to audit: a formal problem statement, assumptions or setup, source/context grounding, ambiguity flags, an evaluation plan, and risks or caveats.

SGHA includes all of these fields for all 15 final project families. This is expected, since the final SGHA report is designed to expose the problem structure rather than only give a proposal title and motivation. The baseline systems produce different kinds of artifacts. AI-Scientist-v2 outputs usually include readable motivation, source/context information, evaluation plans, and caveats, but they do not produce SGHA-style formal problem objects or ambiguity flags. MOOSE-Star provides source-grounded hypothesis text in this setup, but not the full structured research-problem object.

% \begin{table*}[t]
% \centering
% \small
% \caption{Structural artifact coverage across the 15 retained candidates for each method. A checkmark means the field is present for all retained candidates from that method; a cross means it is absent from all retained candidates. This table records artifact structure, not judge-assigned quality.}
% \label{tab:structural-artifact-coverage}
% \resizebox{\textwidth}{!}{
% \begin{tabular}{lccccccc}
% \toprule
% Method & Formal problem & Setup / assumptions & Source/context & Ambiguity flags & Eval. plan & Risks/caveats & Complete artifact \\
% \midrule
% SGHA & \cmark & \cmark & \cmark & \cmark & \cmark & \cmark & \cmark \\
% AI-Scientist-v2 + Qwen & \xmark & \xmark & \cmark & \xmark & \cmark & \cmark & \xmark \\
% AI-Scientist-v2 + Claude Opus & \xmark & \xmark & \cmark & \xmark & \cmark & \cmark & \xmark \\
% MOOSE-Star & \xmark & \xmark & \cmark & \xmark & \xmark & \xmark & \xmark \\
% \bottomrule
% \end{tabular}
% }
% \end{table*}

\begin{table*}[t]
\centering
\small
\caption{Structural artifact coverage across the 15 retained candidates for each method. A checkmark means the field is present for all retained candidates from that method; a cross means it is absent from all retained candidates. This table records artifact structure, not judge-assigned quality.}
\label{tab:structural-artifact-coverage}
\begingroup
\renewcommand{\arraystretch}{1.18}
\resizebox{\textwidth}{!}{
\begin{tabular}{L{0.24\linewidth}ccccccc}
\toprule
\tableheader
\theadcell{Method} & \theadcell{Formal problem} & \theadcell{Setup / assumptions} & \theadcell{Source/context} & \theadcell{Ambiguity flags} & \theadcell{Eval. plan} & \theadcell{Risks/caveats} & \theadcell{Complete artifact} \\
\midrule
\textbf{SGHA} & \yesfield & \yesfield & \yesfield & \yesfield & \yesfield & \yesfield & \yesfield \\
AI-Scientist-v2 + Qwen & \nofield & \nofield & \yesfield & \nofield & \yesfield & \yesfield & \nofield \\
AI-Scientist-v2 + Claude Opus & \nofield & \nofield & \yesfield & \nofield & \yesfield & \yesfield & \nofield \\
MOOSE-Star & \nofield & \nofield & \yesfield & \nofield & \nofield & \nofield & \nofield \\
\bottomrule
\end{tabular}
}
\endgroup
\end{table*}

This analysis helps explain the pattern in Table~\ref{tab:formulation-quality-results}. SGHA's advantage is strongest on source-grounded specificity and ambiguity hygiene because these fields are part of the final artifact. The baselines can still produce clear or well-scoped ideas, especially with a stronger generator, but their outputs are less explicit about the assumptions, missing definitions, and formal structure that a researcher would need to inspect before developing the idea further.

\subsection{Qualitative Comparison of Generated Artifacts}
\label{sec:qualitative-results}

The quantitative results compare average formulation quality across methods. To make the comparison more concrete, we also show representative generated formulations. For each method, we display its best retained Bandits candidate under the formulation-quality evaluation. This keeps the domain fixed while showing the different forms of output produced by SGHA, AI-Scientist-v2, and MOOSE-Star.

Table~\ref{tab:qualitative-example-overview} summarizes the four examples. The examples are shown in their generated form: SGHA produces a formalized project-family artifact, AI-Scientist-v2 produces proposal-style ideas, and MOOSE-Star produces a compact source-grounded hypothesis.

% \begin{table*}[t]
% \centering
% \small
% \caption{Representative Bandits artifacts shown in this section. Each row uses the best retained Bandits candidate for that method under the formulation-quality evaluation.}
% \label{tab:qualitative-example-overview}
% \begin{tabular}{p{0.19\linewidth}p{0.42\linewidth}p{0.12\linewidth}p{0.18\linewidth}}
% \toprule
% Method & Example title & Overall & Artifact type \\
% \midrule
% SGHA &
% Characterizing Identifiability Limits of Structured Bandits Under Piecewise Non-Stationarity &
% 6.6 &
% Formalized project family \\

% AI-Scientist-v2 + Qwen &
% Bandits with Adversarial Arm Execution: Robust Learning When Your Choices Don't Match Reality &
% 4.6 &
% Proposal-style idea \\

% AI-Scientist-v2 + Claude Opus &
% Whose Scale Is It Anyway? Monotone-Robust Bandits with Utility Elicitation &
% 6.2 &
% Detailed proposal-style idea \\

% MOOSE-Star &
% Order Preservation in Set-Size Dependent Combinatorial Bandits &
% 2.2 &
% Source-grounded hypothesis \\
% \bottomrule
% \end{tabular}
% \end{table*}

\begin{table*}[t]
\centering
\small
\caption{Representative Bandits artifacts shown in this section. Each row uses the best retained Bandits candidate for that method under the formulation-quality evaluation.}
\label{tab:qualitative-example-overview}
\begingroup
\renewcommand{\arraystretch}{1.16}
\begin{tabular}{L{0.20\linewidth}L{0.43\linewidth}C{0.10\linewidth}L{0.17\linewidth}}
\toprule
\tableheader
\theadcell{Method} & \theadcell{Example title} & \theadcell{Overall} & \theadcell{Artifact type} \\
\midrule
\textbf{SGHA} &
Characterizing Identifiability Limits of Structured Bandits Under Piecewise Non-Stationarity &
\textbf{6.6} &
Formalized project family \\

\addlinespace[0.25em]
\textbf{AI-Scientist-v2 + Qwen} &
Bandits with Adversarial Arm Execution: Robust Learning When Your Choices Don't Match Reality &
4.6 &
Proposal-style idea \\

\addlinespace[0.25em]
\textbf{AI-Scientist-v2 + Claude Opus} &
Whose Scale Is It Anyway? Monotone-Robust Bandits with Utility Elicitation &
6.2 &
Detailed proposal-style idea \\

\addlinespace[0.25em]
\textbf{MOOSE-Star} &
Order Preservation in Set-Size Dependent Combinatorial Bandits &
2.2 &
Source-grounded hypothesis \\
\bottomrule
\end{tabular}
\endgroup
\end{table*}

\subsubsection{SGHA}

\begin{artifactbox}{Characterizing Identifiability Limits of Structured Bandits Under Piecewise Non-Stationarity}

\artifactfield{Problem formulation}
Current robust bandit algorithms for structured action sets, such as Network Lasso, rely on the i.i.d. assumption of context generation. This assumption fails in environments with piecewise constant non-stationarity, leading to unbounded regret and failure in identifying optimal arms. The fundamental question is not merely how to fix a specific algorithm, but whether the network structure itself allows for consistent learning under such distributional shifts, or if a fundamental identifiability barrier exists.

\artifactfield{Proposal-style abstract}
This project studies the fundamental limits of learning in contextual bandits with network-structured action sets when the underlying data distribution undergoes piecewise constant shifts. The central question is whether the structural constraints imposed by network regularization are sufficient to guarantee consistent policy identification in non-stationary regimes, or if the combination of structural sparsity and temporal drift creates an inherent identifiability gap. A successful outcome would characterize the precise boundary between learnable and unlearnable regimes for this problem class, providing necessary and sufficient conditions for robustness that are independent of any specific algorithmic implementation. This work moves beyond validating a single method to establishing a theoretical framework for the viability of structured learning under non-stationarity.

\artifactfield{Formal problem statement}
Let \(\mathcal{A}\) be a set of actions constrained by a network structure \(\mathcal{G}\). Let \(\mathcal{D}_t\) denote the distribution of contexts at time \(t\). The environment exhibits piecewise constant non-stationarity, meaning \(\mathcal{D}_t = \mathcal{D}_k\) for \(t \in [t_k,t_{k+1})\). The question is whether there exists a sequence of policies \(\pi_t\) such that the regret \(R_T\) grows sublinearly with time \(T\) solely based on the structural constraints of \(\mathcal{G}\), without assuming that \(\mathcal{D}_t\) is i.i.d. across time.

\artifactfield{Variables and notation}
\(\mathcal{A}\): set of available actions;
\(\mathcal{G}\): network structure constraining action sets;
\(\mathcal{D}_t\): distribution of contexts at time \(t\);
\(t_k\): time points where the distribution shifts;
\(\pi_t\): policy at time \(t\);
\(R_T\): cumulative regret up to time \(T\).

\artifactfield{Objective}
Characterize the necessary and sufficient conditions for consistent learning, or identifiability, in network-structured bandits under piecewise constant non-stationarity.

\artifactfield{Assumptions}
Network structure constraint: kept.
Piecewise constant non-stationarity: kept.
i.i.d. context generation: removed.
Existence of consistent learning: questioned.

\artifactfield{Success criterion}
Establish a precise boundary between regimes where network structure aids robustness and regimes where structural constraints amplify non-stationarity errors, leading to impossibility.

\artifactfield{Risk and falsification condition}
Proving impossibility results requires rigorous mathematical machinery, and the result may depend on how the structural and non-stationarity assumptions are defined. The direction would be weakened by a specific non-stationary environment and network topology where consistent learning is achieved despite violating the proposed necessary conditions for failure.

\artifactfield{Ambiguity flags}
Network structure; piecewise constant non-stationarity; feedback model; boundary; feedback or measurement model.

\artifactfield{Source/context grounding}
The family traces to a verification-passed structural gap, \texttt{gap:ba210076fbccfacb}, with source direct formulation \texttt{direct:02}. The retained supporting papers are \emph{Clusters Agnostic Network Lasso Bandits} and \emph{Network Lasso Bandits}. In the SGHA trace, these papers support an assumption-mismatch pattern around Network Lasso-style bandit methods: the method is linked to i.i.d. context/action-set assumptions and to a piecewise-constant non-stationarity failure condition.

\end{artifactbox}

This SGHA example is a structured research-problem object. It contains a problem formulation, an abstract, notation, an objective, assumption statuses, a success criterion, a risk, a falsification condition, ambiguity flags, and source grounding. The artifact is still a candidate direction, but the information needed to inspect and refine it is made explicit.

\subsubsection{AI-Scientist-v2 with Qwen}

\begin{artifactbox}{Bandits with Adversarial Arm Execution}

\artifactfield{Problem statement}
Standard bandit algorithms assume perfect action execution, but in real systems such as cloud services, autonomous systems, and recommendation platforms, there is often uncertainty between the learner's intended action and the action actually executed because of system errors, competing processes, or adversarial interference. The hypothesis is that explicitly modeling execution uncertainty, where an adversary or system can alter which arm is actually executed, requires fundamentally different algorithmic approaches that can achieve robust regret bounds even when the learner observes only the outcome and not the discrepancy.

\artifactfield{Motivation}
In practical bandit deployments, the gap between intended action and actual execution can severely undermine learning. System failures, resource contention, or adversarial interference may cause the learner's selected arm to differ from the executed arm, yet standard bandit algorithms operate under the assumption of perfect execution. This creates a critical vulnerability: learners may systematically misattribute rewards to wrong arms, leading to catastrophic regret growth.

\artifactfield{Proposed direction}
The proposed Adversarial Arm Execution bandit framework allows an adversary or stochastic system to alter the executed arm after the learner's selection, with the learner observing only the final executed arm and its reward.

\artifactfield{Evaluation plan}
The proposed evaluation includes a synthetic adversarial-arm-execution benchmark where the executed arm differs from the selected arm in a fraction of trials; comparisons between adversarial and stochastic execution uncertainty; a recovery analysis where occasional ground-truth signals reveal the executed arm identity; and a real-world simulation using recommendation logs where impressions may not match clicks because of rendering errors.

\artifactfield{Risks}
The generated artifact notes computational complexity, modeling assumptions, limited ground truth, and adversarial complexity as risks.

\end{artifactbox}

The Qwen AI-Scientist-v2 example reads as a proposal for a new bandit setting. It gives a clear motivation, describes a setting, and outlines experiments. Its emphasis is on a project direction and evaluation plan rather than on a formalized problem object.

\subsubsection{AI-Scientist-v2 with Claude Opus}

\begin{artifactbox}{Whose Scale Is It Anyway? Monotone-Robust Bandits with Utility Elicitation}

\artifactfield{Problem statement}
Almost every deployed bandit optimizes the mean of an arbitrarily chosen numerical encoding of an ordinal signal, such as 1--5 stars, 1--10 judge scores, or clinical grades. The identity of the best arm is not invariant to monotone re-encodings of that signal. The proposed direction hypothesizes that such rank reversals and even non-transitive dominance cycles are common in logged feedback, that no algorithm using only sampled rewards can fully resolve them, and that a scale-free learner should play a randomized minimax-regret mixture over arms unless a small number of utility-elicitation queries is available.

\artifactfield{Motivation}
Bandit algorithms maximize the mean of a number, but in many applications that number is an arbitrary encoding of an ordinal signal. A five-star rating, a 1--10 LLM-judge score, or a clinical grade can be mapped to utilities in multiple monotone ways, and the choice can flip which arm is best.

\artifactfield{Proposed direction}
The artifact formalizes a monotone-robust bandit setting in which rewards take values in an ordered set of \(L\) levels, the true utility \(u\) is an unknown increasing function, and the learner is judged by regret maximized over all utilities consistent with what it knows. It proposes studying rank reversals and minimax regret over monotone utility functions, together with utility-elicitation queries based on standard-gamble comparisons over the \(L\) observation levels.

\artifactfield{Evaluation plan}
The proposed evaluation includes scale-fragility audits on rating datasets, synthetic ordinal bandits, randomization-gap experiments, elicitation-complexity experiments, a joint exploration-elicitation tradeoff, and an LLM-evaluation case study.

\artifactfield{Risks}
The generated artifact notes that the novelty depends on the minimax-regret-over-utilities objective, the randomization result, and the elicitation-complexity result. It also notes that unconstrained monotone utilities may make non-comparability pervasive, that constants may be conservative, and that logged ratings are confounded.

\end{artifactbox}

The Claude Opus AI-Scientist-v2 example is the strongest baseline example in this qualitative set. It is detailed, technically specific, and well motivated. It also helps explain why this baseline scores strongly on clarity, technical specificity, nontriviality, and scope control.

\subsubsection{MOOSE-Star}

\begin{artifactbox}{Order Preservation in Set-Size Dependent Combinatorial Bandits}

\artifactfield{Hypothesis}
The order preservation property addresses the gap in handling larger exploration sets by maintaining the order of reward means, which allows for a more efficient exploration strategy. This reduces the regret associated with traditional methods that struggle with extensive exploration sets.

\artifactfield{Mechanism}
The property ensures that the order of reward means remains consistent regardless of set size, enabling the algorithm to focus on the most promising combinations of base arms. This prioritization allows for more efficient exploration of superarms, reducing the number of necessary trials and thus lowering regret.

\artifactfield{Proposed direction}
The SUCB algorithm is adapted to incorporate the order preservation property by modifying the selection process to prioritize superarms based on the order of their base arms' rewards.

\artifactfield{Expected contribution}
The expected contribution is to use order preservation to focus exploration on high-potential combinations of base arms, reducing unnecessary trials and lowering regret.

\artifactfield{Source grounding}
Inspiration paper: \emph{Set-Size Dependent Combinatorial Bandits}.

\end{artifactbox}

The MOOSE-Star example is a compact hypothesis-style artifact. It is grounded in an inspiration paper and proposes a mechanism, but it exposes fewer fields for inspection than the SGHA or AI-Scientist-v2 examples.

\subsubsection{Summary}

The qualitative examples clarify what the numerical scores summarize. SGHA returns a formalized project-family artifact, with notation, assumptions, a success criterion, risks, ambiguity flags, and source grounding. AI-Scientist-v2 returns proposal-style ideas, which can be readable and technically detailed, especially with a stronger generator. MOOSE-Star returns a shorter source-grounded hypothesis. These differences in artifact shape align with the quantitative results: SGHA is strongest on source-grounded specificity, formalizability, and ambiguity hygiene, while the strongest ideation baseline is competitive on clarity, technical specificity, and scope control.

\subsection{Profile-Conditioned Generation}
\label{sec:profile-results}

We next evaluate whether SGHA can be steered by a researcher-specific literature context. In this setting, the input includes a profile-derived corpus and, optionally, a target topic. The downstream SGHA pipeline remains the same: the system still extracts evidence, builds a graph, detects candidate gaps, verifies them, formulates them, applies the critic, and produces final project families. Thus, profile conditioning changes where the evidence comes from.

Table~\ref{tab:profile-pipeline-results} reports the pipeline yield for three profile-conditioned runs. Across the three profiles, SGHA extracts structured content from 439 papers and produces 3,404 evidence tuples. Nine gaps pass verification, leading to four final project families with four formal problem statements.

% \begin{table*}[t]
% \centering
% \small
% \caption{Profile-conditioned SGHA pipeline results. Verified gaps are candidate gaps that pass the hard verification gate before formulation.}
% \label{tab:profile-pipeline-results}
% \begin{tabular}{lrrrrrrr}
% \toprule
% Profile & Seeds & Parsed & Extracted & Tuples & Verified gaps & Families & Formal statements \\
% \midrule
% Yann LeCun & 11 & 150 & 148 & 1170 & 4 & 1 & 1 \\
% Geoffrey Hinton & 33 & 150 & 141 & 1191 & 1 & 1 & 1 \\
% Michael I. Jordan & 150 & 150 & 150 & 1043 & 4 & 2 & 2 \\
% \midrule
% Total & 194 & 450 & 439 & 3404 & 9 & 4 & 4 \\
% \bottomrule
% \end{tabular}
% \end{table*}
\begin{table*}[t]
\centering
\small
\caption{Profile-conditioned SGHA pipeline results. Verified gaps are candidate gaps that pass the hard verification gate before formulation.}
\label{tab:profile-pipeline-results}
\begingroup
\renewcommand{\arraystretch}{1.12}
\begin{tabular}{L{0.24\linewidth}rrrrrrr}
\toprule
\tableheader
\theadcell{Profile} & \theadcell{Seeds} & \theadcell{Parsed} & \theadcell{Extracted} & \theadcell{Tuples} & \theadcell{Verified gaps} & \theadcell{Families} & \theadcell{Formal statements} \\
\midrule
\textbf{Yann LeCun} & 11 & 150 & 148 & 1170 & 4 & 1 & 1 \\
\textbf{Geoffrey Hinton} & 33 & 150 & 141 & 1191 & 1 & 1 & 1 \\
\textbf{Michael I. Jordan} & 150 & 150 & 150 & 1043 & 4 & 2 & 2 \\
\midrule
\tabtotal
\textbf{Total} & \textbf{194} & \textbf{450} & \textbf{439} & \textbf{3404} & \textbf{9} & \textbf{4} & \textbf{4} \\
\bottomrule
\end{tabular}
\endgroup
\end{table*}

We evaluate the resulting candidates with a personalized judge rubric. This rubric keeps the formulation-quality criteria from the main evaluation and adds profile-specific criteria: profile alignment, profile specificity, intellectual-style match, and personalization overall. The judge uses only the provided profile context and candidate text.

Table~\ref{tab:profile-score-results} shows the personalized scores. The strongest profile-alignment scores appear for the Geoffrey Hinton candidate and the Michael I. Jordan hierarchical-identifiability candidate. The Yann LeCun candidate has the highest formulation-quality score among the profile-conditioned outputs, but lower profile specificity. This pattern is useful: profile conditioning can steer SGHA toward profile-relevant directions, while the judge still separates general formulation quality from how specifically the candidate matches the provided profile context.

% \begin{table*}[t]
% \centering
% \scriptsize
% \caption{Profile-conditioned judge scores. Formulation measures the research-problem formulation itself; the remaining columns measure alignment to the provided profile context.}
% \label{tab:profile-score-results}
% \resizebox{\textwidth}{!}{
% \begin{tabular}{lp{0.45\textwidth}rrrrr}
% \toprule
% Profile & Candidate & Formulation & Alignment & Specificity & Style match & Personalization \\
% \midrule
% Yann LeCun &
% Fundamental limits of partition-function approximation in high-dimensional EBMs &
% 6.17 & 7.00 & 4.50 & 5.67 & 5.33 \\

% Geoffrey Hinton &
% Convergence failure regimes of stochastic-gradient approximations in deep generative models &
% 5.67 & 8.00 & 7.17 & 6.83 & 7.17 \\

% Michael I. Jordan &
% Identifiability boundaries of filtering-clustering under rank-deficient design matrices &
% 6.00 & 6.50 & 5.83 & 7.33 & 6.33 \\

% Michael I. Jordan &
% Characterizing hierarchical identifiability limits in variational protein annotation &
% 6.00 & 8.00 & 7.00 & 7.75 & 7.00 \\
% \bottomrule
% \end{tabular}
% }
% \end{table*}

\begin{table*}[t]
\centering
\scriptsize
\caption{Profile-conditioned judge scores. Formulation measures the research-problem formulation itself; the remaining columns measure alignment to the provided profile context.}
\label{tab:profile-score-results}
\begingroup
\renewcommand{\arraystretch}{1.14}
\resizebox{\textwidth}{!}{
\begin{tabular}{L{0.18\linewidth}L{0.43\linewidth}rrrrr}
\toprule
\tableheader
\theadcell{Profile} & \theadcell{Candidate} & \theadcell{Formulation} & \theadcell{Alignment} & \theadcell{Specificity} & \theadcell{Style match} & \theadcell{Personalization} \\
\midrule
\textbf{Yann LeCun} &
Fundamental limits of partition-function approximation in high-dimensional EBMs &
6.17 & 7.00 & 4.50 & 5.67 & 5.33 \\

\textbf{Geoffrey Hinton} &
Convergence failure regimes of stochastic-gradient approximations in deep generative models &
5.67 & \bestscore{8.00} & 7.17 & 6.83 & \bestscore{7.17} \\

\textbf{Michael I. Jordan} &
Identifiability boundaries of filtering-clustering under rank-deficient design matrices &
6.00 & 6.50 & 5.83 & 7.33 & 6.33 \\

\textbf{Michael I. Jordan} &
Characterizing hierarchical identifiability limits in variational protein annotation &
6.00 & \bestscore{8.00} & 7.00 & \bestscore{7.75} & 7.00 \\
\bottomrule
\end{tabular}
}
\endgroup
\end{table*}

The Michael I. Jordan example illustrates the kind of profile-conditioned artifact SGHA produces. The final family, \emph{Characterizing Hierarchical Identifiability Limits in Variational Protein Annotation}, connects profile-relevant themes such as variational inference, graphical models, and structured prediction to a concrete problem in protein-function annotation.

\begin{artifactbox}{Characterizing Hierarchical Identifiability Limits in Variational Protein Annotation}

\artifactfield{Profile context}
This candidate comes from the Michael I. Jordan profile-conditioned run. The profile-conditioned evidence base emphasizes probabilistic modeling, graphical models, statistical machine learning, variational inference, Bayesian nonparametrics, and optimization. The run uses a profile-conditioned corpus of 150 parsed and extracted papers, producing 1,043 evidence tuples.

\artifactfield{Source papers}
The final family is grounded in the retained profile-seed papers profile pdf: a variational principle for graphical models \cite{wainwright200511} and profile pdf: genome scale phylogenetic function annotation of large and diverse protein families \cite{engelhardt2011genome}.
The first source provides the variational-inference / graphical-model context, while the second provides the protein-function annotation setting.

\artifactfield{Verified gap}
The family traces to the verified gap \texttt{gap:dcb6662b99312fb7}, an unresolved tradeoff around Gene Ontology hierarchy. The gap passes the hard verification gate with survival score \(0.664\), above the threshold \(0.600\). The verification trace is mixed in a useful way: the support and feasibility agents view the gap as grounded and studyable, while the skeptic and critic caution that the strongest version of the claim should not be overstated because hierarchy-aware scoring methods exist.

\artifactfield{Direct formulation}
The source direct formulation is \texttt{direct:04}: \emph{Integrating Gene Ontology Hierarchy Constraints into Variational Inference for Protein Function Annotation}. This direct formulation asks for a variational inference framework that explicitly incorporates Gene Ontology tree constraints while avoiding prohibitive computational cost.

\artifactfield{Ambition expansion and critic}
The ambition stage generates three variants. The first two variants, \texttt{var:10} and \texttt{var:11}, are rejected by the independent critic as inflated or insufficiently supported. The accepted variant is \texttt{var:12}, which reframes the problem as a theoretical boundary: under what structural conditions do flat variational objectives fail to recover the true posterior over a tree-structured label space? This accepted variant becomes the final project family.

\artifactfield{Problem formulation}
Current variational models for protein function annotation treat Gene Ontology (GO) terms as flat sets, ignoring the parent-child hierarchy. This oversight creates a fundamental gap where biologically consistent, less precise predictions are penalized, but the theoretical boundary of this failure is unknown. The central challenge is to characterize the exact conditions under which hierarchical constraints become necessary for identifiability and to define the regime where flat approximations fail catastrophically.

\artifactfield{Formal problem statement}
Let \(\mathcal{T}\) be a tree-structured label space representing Gene Ontology terms and parent-child relationships. Let \(P_{\mathrm{true}}\) be the true posterior over \(\mathcal{T}\) given protein data \(D\), and let \(\mathcal{F}_{\mathrm{flat}}\) be a flat variational approximation family. Each \(Q \in \mathcal{F}_{\mathrm{flat}}\) approximates \(P_{\mathrm{true}}\) without enforcing parent-child consistency constraints from the tree topology. The problem is to characterize conditions \(\mathcal{C}\) such that, for every \(Q \in \mathcal{F}_{\mathrm{flat}}\), the approximation error \(\|P_{\mathrm{true}} - Q\|\) exceeds a threshold \(\epsilon\) when the structural disconnect between parent-child implications prevents recovery of the true posterior.

\artifactfield{Variables and notation}
\(\mathcal{T}\): tree-structured Gene Ontology label space;
\(P_{\mathrm{true}}\): hierarchy-respecting posterior over GO terms;
\(\mathcal{F}_{\mathrm{flat}}\): flat variational family;
\(Q\): approximating distribution in \(\mathcal{F}_{\mathrm{flat}}\);
\(D\): observed protein data;
\(\epsilon\): approximation-error threshold;
\(\mathcal{C}\): structural or data conditions under which flat approximations fail.

\artifactfield{Objective}
Characterize when flat variational approximations are insufficient for recovering a hierarchy-respecting posterior over Gene Ontology terms. More concretely, identify structural conditions on the label hierarchy and data distribution under which every flat approximation \(Q \in \mathcal{F}_{\mathrm{flat}}\) remains farther than \(\epsilon\) from \(P_{\mathrm{true}}\).

\artifactfield{Assumptions}
Existence of a hierarchy-respecting posterior: kept.
Gene Ontology tree structure: kept.
Flat variational sufficiency: relaxed or questioned.
Parent-child consistency constraints: treated as necessary to define the hierarchical setting.

\artifactfield{Success criterion}
Identify conditions on the GO tree and data distribution under which flat variational inference fails to recover the hierarchy-respecting posterior, and characterize the resulting identifiability boundary.

\artifactfield{Possible result types}
A theorem characterizing when flat variational families cannot approximate the true hierarchy-respecting posterior within threshold \(\epsilon\); an algorithmic direction for hierarchy-aware variational inference; and an empirical protocol comparing flat and hierarchy-aware approximations on protein-function annotation tasks.

\artifactfield{Ambiguity flags}
\texttt{Catastrophic failure}: requires a specific divergence metric or threshold.
\texttt{Structural disconnect}: requires a precise mathematical relationship between ignored parent-child constraints and approximation error.
\texttt{Boundary}: requires a formal definition of the transition between identifiable and non-identifiable regimes.
\texttt{Feedback or measurement model}: requires further definition for the protein-annotation setting.

\artifactfield{Formalization confidence and risk}
The formalization confidence is \texttt{medium}, and the artifact sets \texttt{requires\_human\_definition} to true. The main risk is that the impossibility-boundary claim is only moderately supported by the source evidence: the sources identify the hierarchy-related gap and the variational-inference context, but do not themselves prove a theoretical failure regime.

\end{artifactbox}
This example shows the intended role of profile conditioning. The profile context changes the evidence base and helps steer SGHA toward a problem that fits the supplied research context, while the final artifact still exposes the formal setup, objective, and remaining ambiguities.

\subsection{Evolutionary Exploration}
\label{sec:evolutionary-results}

We next study SGHA's evolutionary exploration branch. This branch has a different role from the main project-family path. The main path is designed to produce a compact set of verified, critic-filtered, and formalized problem families. Evolutionary exploration instead searches more broadly around literature-grounded seeds. Starting from a seed gap or formulation, it varies parts of the problem representation---such as assumptions, mechanisms, objectives, or regimes---and then critiques, selects, and refines the resulting variants.

Table~\ref{tab:evolutionary-artifact-counts} shows the scale of this exploration. Across the five domains, the branch generates 2,908 evolved candidates and retains 160 selected hypotheses during the search. The final ranked files contain 96 hypotheses. This gives SGHA a broader exploratory layer: the family-report path produces a small set of polished problem families, while the evolutionary branch gives a larger pool of nearby directions that can be inspected or passed through later verification and formalization.

% \begin{table}[t]
% \centering
% \small
% \caption{Evolutionary exploration artifacts by domain. Evolved candidates include mutations, crossovers, and syntheses around literature-grounded seeds.}
% \label{tab:evolutionary-artifact-counts}
% \begin{tabular}{lrrrr}
% \toprule
% Domain & Evolved & Selected & Final ranked & Initial gaps \\
% \midrule
% Bandits & 603 & 32 & 20 & 13 \\
% In-context learning & 472 & 32 & 20 & 16 \\
% Reasoning / test-time computation & 502 & 32 & 20 & 12 \\
% Offline RL & 594 & 32 & 16 & 8 \\
% Uncertainty estimation & 737 & 32 & 20 & 15 \\
% \midrule
% Total & 2908 & 160 & 96 & 64 \\
% \bottomrule
% \end{tabular}
% \end{table}

\begin{table}[t]
\centering
\small
\caption{Evolutionary exploration artifacts by domain. Evolved candidates include mutations, crossovers, and syntheses around literature-grounded seeds.}
\label{tab:evolutionary-artifact-counts}
\begingroup
\renewcommand{\arraystretch}{1.12}
\begin{tabular}{L{0.34\linewidth}rrrr}
\toprule
\tableheader
\theadcell{Domain} & \theadcell{Evolved} & \theadcell{Selected} & \theadcell{Final ranked} & \theadcell{Initial gaps} \\
\midrule
Bandits & 603 & 32 & 20 & 13 \\
In-context learning & 472 & 32 & 20 & 16 \\
Reasoning / test-time computation & 502 & 32 & 20 & 12 \\
Offline RL & 594 & 32 & 16 & 8 \\
Uncertainty estimation & 737 & 32 & 20 & 15 \\
\midrule
\tabtotal
\textbf{Total} & \textbf{2908} & \textbf{160} & \textbf{96} & \textbf{64} \\
\bottomrule
\end{tabular}
\endgroup
\end{table}

We evaluate an output-count-matched set of 15 ranked evolutionary hypotheses with the same formulation-only rubric used in the main comparison. Table~\ref{tab:evolutionary-scores} reports the average scores. The evolutionary hypotheses score highest on source-grounded specificity and nontriviality, which is consistent with their purpose: they are generated from literature-grounded seeds and often point to interesting directions. Their lower scores on well-posedness, formalizability, and ambiguity hygiene also match the role of this branch. These hypotheses are exploratory; they have not gone through the full family-report path that adds formal problem statements, assumption tables, and ambiguity flags.

% \begin{table*}[t]
% \centering
% \small
% \caption{Formulation-only scores for the selected evolutionary-exploration hypotheses. Scores are averaged over the available judge evaluations.}
% \label{tab:evolutionary-scores}
% \resizebox{\textwidth}{!}{
% \begin{tabular}{lrrrrrrrrrr}
% \toprule
% Candidate set & Overall & Clarity & Technical & Well-posed & Boundary & Formal. & Nontriv. & Scope & Grounding & Ambiguity \\
% \midrule
% SGHA evolutionary exploration &
% 3.82 & 4.11 & 4.09 & 2.80 & 3.42 & 2.64 & 5.05 & 4.38 & 5.40 & 2.20 \\
% \bottomrule
% \end{tabular}
% }
% \end{table*}

\begin{table*}[t]
\centering
\small
\caption{Formulation-only scores for the selected evolutionary-exploration hypotheses. Scores are averaged over the available judge evaluations.}
\label{tab:evolutionary-scores}
\begingroup
\renewcommand{\arraystretch}{1.14}
\resizebox{\textwidth}{!}{
\begin{tabular}{L{0.27\linewidth}rrrrrrrrrr}
\toprule
\tableheader
\theadcell{Candidate set} & \theadcell{Overall} & \theadcell{Clarity} & \theadcell{Technical} & \theadcell{Well-posed} & \theadcell{Boundary} & \theadcell{Formal.} & \theadcell{Nontriv.} & \theadcell{Scope} & \theadcell{Grounding} & \theadcell{Ambiguity} \\
\midrule
SGHA evolutionary exploration &
3.82 & 4.11 & 4.09 & 2.80 & 3.42 & 2.64 & 5.05 & 4.38 & 5.40 & 2.20 \\
\bottomrule
\end{tabular}
}
\endgroup
\end{table*}

The domain-level results in Table~\ref{tab:evolutionary-domain-scores} show the same pattern. The reasoning/test-time-computation candidate has the highest mean overall score, and source grounding is relatively strong across domains. This suggests that the evolutionary branch is useful for surfacing grounded directions, even when the resulting hypotheses still need the later formalization and ambiguity-cleanup stages before they become final research-problem artifacts.

% \begin{table}[t]
% \centering
% \small
% \caption{Evolutionary-exploration formulation scores by domain.}
% \label{tab:evolutionary-domain-scores}
% \begin{tabular}{lrrrr}
% \toprule
% Domain & Selected & Mean overall & Best overall & Mean grounding \\
% \midrule
% Bandits & 3 & 3.75 & 4.25 & 5.17 \\
% In-context learning & 4 & 3.64 & 4.33 & 5.29 \\
% Reasoning / test-time computation & 1 & 4.50 & 4.50 & 5.75 \\
% Offline RL & 1 & 3.75 & 3.75 & 5.25 \\
% Uncertainty estimation & 6 & 3.86 & 4.00 & 5.57 \\
% \bottomrule
% \end{tabular}
% \end{table}
\begin{table}[t]
\centering
\small
\caption{Evolutionary-exploration formulation scores by domain.}
\label{tab:evolutionary-domain-scores}
\begingroup
\renewcommand{\arraystretch}{1.12}
\begin{tabular}{L{0.34\linewidth}rrrr}
\toprule
\tableheader
\theadcell{Domain} & \theadcell{Selected} & \theadcell{Mean overall} & \theadcell{Best overall} & \theadcell{Mean grounding} \\
\midrule
Bandits & 3 & 3.75 & 4.25 & 5.17 \\
In-context learning & 4 & 3.64 & 4.33 & 5.29 \\
Reasoning / test-time computation & 1 & \bestscore{4.50} & \bestscore{4.50} & \bestscore{5.75} \\
Offline RL & 1 & 3.75 & 3.75 & 5.25 \\
Uncertainty estimation & 6 & 3.86 & 4.00 & 5.57 \\
\bottomrule
\end{tabular}
\endgroup
\end{table}

The highest-scoring evolutionary candidate comes from reasoning and test-time computation. It studies self-verification mechanisms in test-time compute scaling, where additional inference-time search may stop improving accuracy when intermediate reasoning steps are difficult to verify.

\begin{artifactbox}{Self-verification mechanisms within test-time compute scaling frameworks}

\artifactfield{Problem statement}
The problem of understanding self-verification mechanisms within test-time compute scaling frameworks that achieve accuracy saturation despite increased inference-time search depth due to the collapse of verifiable correctness assumptions on inherently unverifiable intermediate steps.

\artifactfield{Gap}
The problem of verification-dependent error accumulation that achieves accuracy saturation despite increased inference-time search depth.

\artifactfield{Target}
Self-verification mechanisms within test-time compute scaling frameworks.

\artifactfield{Scope}
Cross-gap synthesis over assumption-mismatch, repeated-unaddressed-limitation, and unrealistic-assumption motifs, narrowed to the most evidence-dense setting.

\artifactfield{Mechanism}
Self-verification attempts to correct linear error accumulation inherent in Chain-of-Thought structures, but relies on verifiable correctness assumptions that collapse when intermediate steps are inherently unverifiable.

\artifactfield{Proposal-style abstract}
This project studies self-verification mechanisms within test-time compute scaling frameworks. The central question is how to maintain accuracy when verifiable correctness assumptions collapse on inherently unverifiable intermediate steps. Current approaches often focus on complex problem-solving tasks where correctness is easily verified on labeled datasets. However, reliance on task-specific few-shot examples can limit generalizability across diverse tasks. Furthermore, existing methods fail to utilize negative samples for implicit exploration and do not improve on out-of-distribution tasks through experiments. The mechanism involves self-verification attempting to correct linear error accumulation inherent in Chain-of-Thought structures. Yet, this relies on assumptions that collapse when intermediate steps are unverifiable. This project would investigate verification-dependent error accumulation that achieves accuracy saturation despite increased inference-time search depth. A possible approach is refining these mechanisms to handle unverifiable steps without saturation. A successful outcome would provide insights into scaling frameworks. This would clarify the relationship between search depth and verification limits. The proposed contribution type focuses on self-verification mechanisms within test-time compute scaling frameworks. This work aims to address the unrealistic assumption that intermediate steps remain verifiable under increased compute scaling.

\artifactfield{Source grounding}
Supporting papers include \texttt{openreview:0vKokoPKTo}, \texttt{openreview:1OyE9IK0kx}, \texttt{openreview:4Po8d9GAfQ}, \texttt{openreview:70YeidEcYR}, \texttt{openreview:77gQUdQhE7}, \texttt{openreview:JtGPIZpOrz}, \texttt{openreview:P6dwZJpJ4m}, \texttt{openreview:YUYJsHOf3c}, \texttt{openreview:iEdEHPcFeu}, and \texttt{openreview:w6nlcS8Kkn}. The originating motifs were assumption mismatch, repeated unaddressed limitation, and unrealistic assumption.

\artifactfield{Score}
Mean overall formulation quality: \(4.50\)

\end{artifactbox}

These results clarify the role of evolutionary exploration. The branch is good at breadth: it produces many source-grounded hypotheses and can combine signals from several related gaps. The more modest formulation scores show where the main family-report path adds value. Formal problem statements, assumption statuses, risks, and ambiguity flags are not just formatting details; they are the stages that turn a promising hypothesis into an inspectable research-problem artifact. In this sense, the evolutionary branch is best viewed as a generator of candidate directions for further refinement, while the main SGHA path produces the final formalized families used in the primary comparison.

\subsection{Model- and Corpus-Size Sensitivity}
\label{sec:sensitivity-results}

We end the results with two Bandit sensitivity studies. These experiments are not meant to establish a scaling law. Instead, they show how SGHA's later stages respond when we change either the model used inside the pipeline or the amount of literature evidence available to it.

\paragraph{Model size.}
Table~\ref{tab:model-size-sensitivity} compares the main 9B Bandits run with smaller 4B and larger 27B configurations. The 4B run produces a much smaller evidence base, but still reaches verification-passed gaps and direct formulations. The run stops before final family construction because none of the ambition-expanded variants pass the critic. The 9B and 27B runs both reach formalized project families. The 27B run produces more verification-passed gaps than the 9B run, but the critic and consolidation stages reduce them to one final family.

% \begin{table*}[t]
% \centering
% \small
% \caption{Bandits model-size sensitivity. Accepted variants are ambition-expanded formulations that pass the independent critic. Mean overall is reported for scoreable final families.}
% \label{tab:model-size-sensitivity}
% \begin{tabular}{lrrrrrrr}
% \toprule
% Model & Tuples & Verified gaps & Direct & Accepted variants & Families & Formal statements & Mean overall \\
% \midrule
% 4B & 204 & 8 & 8 & 0 & 0 & 0 & -- \\
% 9B & 1554 & 6 & 6 & 4 & 3 & 3 & 5.93 \\
% 27B & 1570 & 19 & 19 & 3 & 1 & 1 & 6.67 \\
% \bottomrule
% \end{tabular}
% \end{table*}
\begin{table*}[t]
\centering
\small
\caption{Bandits model-size sensitivity. Accepted variants are ambition-expanded formulations that pass the independent critic. Mean overall is reported for scoreable final families.}
\label{tab:model-size-sensitivity}
\begingroup
\renewcommand{\arraystretch}{1.12}
\begin{tabular}{L{0.12\linewidth}rrrrrrr}
\toprule
\tableheader
\theadcell{Model} & \theadcell{Tuples} & \theadcell{Verified gaps} & \theadcell{Direct} & \theadcell{Accepted variants} & \theadcell{Families} & \theadcell{Formal statements} & \theadcell{Mean overall} \\
\midrule
4B & 204 & 8 & 8 & 0 & 0 & 0 & -- \\
9B & 1554 & 6 & 6 & 4 & 3 & 3 & 5.93 \\
27B & 1570 & 19 & 19 & 3 & 1 & 1 & \bestscore{6.67} \\
\bottomrule
\end{tabular}
\endgroup
\end{table*}

The model-size study shows that capacity changes both the evidence extracted upstream and the shape of the final funnel. A smaller model can still identify formulation-ready gaps, but the final output depends on whether the generated variants survive criticism. A larger model can surface many more verified gaps, while consolidation can still lead to a small final set. This is consistent with SGHA's design: final families are selected through verification, criticism, and consolidation, rather than by a fixed output quota.

\paragraph{Corpus size.}
Table~\ref{tab:corpus-size-sensitivity} varies the Bandits corpus from 50 to 250 selected papers. The 50-paper run reaches a verified gap and a direct formulation, but does not produce a final family. From 100 papers onward, SGHA reaches the formal problem stage. The 200-paper setting produces the largest number of final families and the highest mean overall score in this diagnostic, while the 250-paper setting remains close and produces a complete set of formalized outputs.

% \begin{table*}[t]
% \centering
% \small
% \caption{Bandits corpus-size sensitivity. The 250-paper setting is the main Bandits run.}
% \label{tab:corpus-size-sensitivity}
% \begin{tabular}{lrrrrrrr}
% \toprule
% Selected papers & Tuples & Verified gaps & Direct & Accepted variants & Families & Formal statements & Mean overall \\
% \midrule
% 50 & 339 & 1 & 1 & 0 & 0 & 0 & -- \\
% 100 & 706 & 4 & 4 & 3 & 2 & 2 & 5.80 \\
% 150 & 1036 & 5 & 5 & 1 & 1 & 1 & 5.20 \\
% 200 & 1392 & 10 & 10 & 6 & 5 & 5 & 6.12 \\
% 250 & 1554 & 6 & 6 & 4 & 3 & 3 & 5.93 \\
% \bottomrule
% \end{tabular}
% \end{table*}
\begin{table*}[t]
\centering
\small
\caption{Bandits corpus-size sensitivity. The 250-paper setting is the main Bandits run.}
\label{tab:corpus-size-sensitivity}
\begingroup
\renewcommand{\arraystretch}{1.12}
\begin{tabular}{L{0.16\linewidth}rrrrrrr}
\toprule
\tableheader
\theadcell{Selected papers} & \theadcell{Tuples} & \theadcell{Verified gaps} & \theadcell{Direct} & \theadcell{Accepted variants} & \theadcell{Families} & \theadcell{Formal statements} & \theadcell{Mean overall} \\
\midrule
50 & 339 & 1 & 1 & 0 & 0 & 0 & -- \\
100 & 706 & 4 & 4 & 3 & 2 & 2 & 5.80 \\
150 & 1036 & 5 & 5 & 1 & 1 & 1 & 5.20 \\
200 & 1392 & 10 & 10 & 6 & 5 & 5 & \bestscore{6.12} \\
250 & 1554 & 6 & 6 & 4 & 3 & 3 & 5.93 \\
\bottomrule
\end{tabular}
\endgroup
\end{table*}

The corpus-size study gives a similar message. A richer corpus gives SGHA more evidence and more opportunities for candidate-gap discovery, but the final output is shaped by which gaps pass verification and which formulations survive the critic. The best result among these Bandits settings appears at 200 papers, while the 250-paper run still produces a strong but smaller final set. We therefore treat corpus size as a sensitivity variable: enough evidence is important, but the structure of that evidence, and how it passes through the later gates, matters just as much.

%% file: files/discussion.tex
\section{Discussion}
\label{sec:discussion}

In this paper, we ask whether research-problem generation can start from patterns in the literature rather than from open-ended ideation. The results suggest that this changes the output in a meaningful way. Strong ideation systems, especially AI-Scientist-v2 with Claude Opus, can produce polished and technically detailed proposals. However, SGHA produces a different kind of artifact: a problem family that keeps track of the evidence, the verified gap, the assumptions being questioned, and the definitions that remain unresolved. This difference is reflected in the scores as SGHA is strongest on source-grounded specificity, assumption-boundary clarity, formalizability, and ambiguity hygiene, while the strongest ideation baseline is competitive on clarity, technical detail, nontriviality, and scope control.

This distinction matters because a useful research problem is not only a well-written idea. A researcher also needs to know where the problem came from, what assumption is being relaxed or challenged, what would count as progress, and which parts of the formulation are still underspecified. The qualitative examples make this visible. The AI-Scientist-v2 outputs often read like proposal sketches, with motivations and evaluation plans, while the MOOSE-Star output is closer to a compact hypothesis grounded in an inspiration paper. The SGHA output is more structured: it includes provenance, notation, assumptions, a success criterion, risks, and ambiguity flags. These fields do not make the problem automatically correct or novel, but they make the candidate easier to inspect and refine.

The profile-conditioned and evolutionary experiments further show two ways to use the same system beyond the main run. Profile conditioning changes the evidence base, so the final problems can move toward a researcher-specific context while still passing through the same verification and formulation stages. On the other hand, evolutionary exploration serves a different purpose: it searches more broadly around literature-grounded seeds. Its outputs are useful for breadth and source-grounded hypothesis generation, while the lower formalizability and ambiguity-hygiene scores show why the final family-report path remains important.

The sensitivity studies further are a useful caution against reading SGHA as a simple scaling pipeline. Larger models and larger corpora can change the evidence graph and the set of candidate gaps, but the final output is shaped by verification, criticism, and consolidation. This is appropriate for a literature-driven system: the goal is not to force a fixed number of ideas, but to return problem families when the evidence supports them. At the same time, the current evaluation is limited to formulation quality. The LLM judges do not establish external novelty, theoretical correctness, or future scientific impact, and the semi-formal statements are not finished theorem statements. SGHA should therefore be viewed as a tool for producing evidence-linked starting points that researchers can inspect, sharpen, and compare against the broader literature.

%% file: files/conclusion.tex
\section{Conclusion}
\label{sec:conclusion}

We presented SGHA, a training-free system for generating research-problem families from scientific literature. The central idea is to use the literature itself as the starting point: papers already contain assumptions, limitations, empirical patterns, and open directions that can inherently suggest useful research problems when considered together. SGHA turns these signals into a structured evidence graph, identifies candidate gaps, verifies them, and formulates the surviving candidates as project families with semi-formal problem statements.

Across five machine-learning domains, SGHA produced 15 formalized project families from 1,250 selected papers and 8,634 extracted evidence tuples. In the formulation-only evaluation, SGHA achieved the highest overall score among the compared methods and was strongest on source-grounded specificity, assumption-boundary clarity, formaliability, and ambiguity. The qualitative examples further show the same pattern: strong ideation baselines can produce polished proposal-style ideas, while SGHA produces structured problem artifacts with provenance, assumptions, objectives, risks, and ambiguity flags.

These results point to a broader role for scientific assistants. Beyond summarizing papers or generating ideas from a topic, they can help organize evidence across papers into clearer research-problem formulations. SGHA takes a step in that direction by making the path from literature evidence to problem statement explicit. This gives researchers a more grounded starting point for reading, refining, and developing new research directions.

%% file: files/limitation.tex
\section{Scope and Intended Use}
\label{sec:limitations}

SGHA is aimed at a particular kind of research-problem generation: problems whose evidence is already partly visible in a scientific corpus. Many useful directions begin this way. A set of papers may rely on the same strong assumption, report related failure modes, leave similar limitations unresolved, or evaluate methods under narrow conditions. SGHA is designed to collect these signals, organize them as evidence, and turn the strongest patterns into clearer problem formulations.

This scope is central to the system. SGHA is not positioned as a replacement for open-ended ideation; it is a complementary mode of ideation grounded in the literature. Open-ended ideation is useful when the goal is to speculate freely, connect distant areas, or invent directions that are not yet reflected in papers. SGHA is useful when the goal is to ask: given the literature we have, what problems are already being suggested by its assumptions, gaps, and tensions? The verification and formalization stages are designed around this setting.

The evaluation follows the same framing. We assess whether SGHA produces clear, source-grounded, formalizable problem artifacts with explicit assumptions and ambiguity flags. We do not use the results as a claim that the generated problems are externally novel or already correct in the broader literature. Instead, the intended output is a structured starting point for researchers: a formulation that preserves why the problem arose, what evidence supports it, and what still needs to be sharpened before the direction becomes a complete research project.

%% file: files/genai_disclosure.tex
\section*{Generative AI Disclosure}

The authors use ChatGPT for language editing and formatting. They also use the Stanford Agentic Reviewer\footnote{\url{https://paperreview.ai/}} and the CMU Paper Reviewer \citep{kim2026limits} to obtain preliminary feedback and improve the paper. Codex was also used to organise the code and prepare the README files. All technical content, experimental design, analyses, and conclusions remain the sole responsibility of the authors.

%% file: appendix/appendixa.tex
\appendix

\section{SGHA Prompts, Output Schemas, and Validation Rules}
\label{app:prompts-and-schemas}

This appendix reports the prompt templates, output contracts, and validation rules used by SGHA. Runtime prompts contain instance-specific content, such as paper text, candidate-gap evidence, verification summaries, or project-family records. We therefore show the fixed prompt instructions with placeholders for dynamic fields, such as \texttt{\{paper\_text\}}, \texttt{\{gap\_json\}}, or \texttt{\{verification\_summary\}}.

SGHA uses language models for bounded extraction, review, and formulation stages. The system uses deterministic code for graph construction, motif detection, hard verification gating, family consolidation, and final report auditing. The main SGHA outputs are generated only after candidate gaps pass screening and verification.

\subsection{Evidence Extraction}

The extraction stage reads a parsed paper and returns a structured paper object together with typed evidence tuples. The prompt asks for extracted scientific content from the given paper, not a summary or template.

\paragraph{Prompt template.}

\begin{PromptBox}
You are a scientific information extractor. Read the paper below and extract structured information as JSON.

IMPORTANT: Do NOT return a schema or template. Return a JSON object with REAL extracted content from this specific paper.

Output keys:
  paper_id
  claims
  limitations
  failure_conditions
  contradictions_or_tensions
  future_work
  methods
  assumptions
  tasks
  datasets
  metrics
  results
  tuples

For each tuple:
  subject        -- specific named entity
  relation       -- MUST be one of: evaluated_on, improves_over, fails_under, assumes,
                   contradicts, uses_dataset, measured_by, limited_by, addresses,
                   not_addressed_by
  object
  evidence_text  -- exact supporting text from the paper
  section
  confidence     -- value in [0, 1]
  claim_type
  polarity
  condition
  task
  dataset
  model
  metric
  direction
  evidence_type
  strength
  source_span
  subject_scope
  resolved_by_paper
  in_related_work

Rules:
- Extract content from the provided paper only.
- Use specific scientific entities, not generic placeholders.
- Evidence text and source spans must be grounded in the parsed paper.
- Do not conflate assumptions with failures.
- Do not invent results, limitations, datasets, or methods.

Paper ID: {paper_id}
Title: {title}

Paper text:
{paper_text}
\end{PromptBox}

\paragraph{Paper-level schema.}

\begin{PromptBox}
PaperExtraction:
  paper_id: string
  claims: list[string]
  limitations: list[string]
  failure_conditions: list[string]
  contradictions_or_tensions: list[string]
  future_work: list[string]
  methods: list[string]
  assumptions: list[string]
  tasks: list[string]
  datasets: list[string]
  metrics: list[string]
  results: list[string]
  tuples: list[ScientificTuple]
\end{PromptBox}

\paragraph{Tuple schema.}

\begin{table*}[t]
\centering
\small
\caption{Evidence tuple schema used by SGHA.}
\label{tab:app-evidence-tuple-schema}
\begingroup
\renewcommand{\arraystretch}{1.12}
\begin{tabular}{L{0.26\linewidth}L{0.64\linewidth}}
\toprule
\tableheader
\theadcell{Field} & \theadcell{Meaning} \\
\midrule
\texttt{subject} & Specific method, model, assumption, task, dataset, metric, claim, or scientific object. \\
\texttt{relation} & One of \texttt{evaluated\_on}, \texttt{improves\_over}, \texttt{fails\_under}, \texttt{assumes}, \texttt{contradicts}, \texttt{uses\_dataset}, \texttt{measured\_by}, \texttt{limited\_by}, \texttt{addresses}, or \texttt{not\_addressed\_by}. \\
\texttt{object} & Target of the relation. \\
\texttt{evidence\_text} / \texttt{source\_span} & Supporting text from the parsed paper. \\
\texttt{section} & Paper section containing the evidence. \\
\texttt{confidence} & Model confidence in the extraction. \\
\texttt{claim\_type} & One of \texttt{method}, \texttt{result}, \texttt{limitation}, \texttt{assumption}, \texttt{comparison}, \texttt{failure}, \texttt{hypothesis}, \texttt{evaluation}, or \texttt{unknown}. \\
\texttt{polarity} & One of \texttt{positive}, \texttt{negative}, \texttt{neutral}, \texttt{mixed}, or \texttt{unknown}. \\
\texttt{condition} & Regime or condition under which the tuple holds, if available. \\
\texttt{task}, \texttt{dataset}, \texttt{model}, \texttt{metric} & Structured context fields, especially for evaluation-related claims. \\
\texttt{direction} & One of \texttt{improves}, \texttt{worsens}, \texttt{no\_change}, or \texttt{unknown}. \\
\texttt{evidence\_type} & One of \texttt{empirical}, \texttt{theoretical}, \texttt{ablation}, \texttt{qualitative}, \texttt{assumption}, \texttt{citation}, \texttt{benchmark}, or \texttt{unknown}. \\
\texttt{strength} & One of \texttt{strong}, \texttt{moderate}, \texttt{weak}, or \texttt{unknown}. \\
\texttt{subject\_scope} & One of \texttt{own\_method}, \texttt{prior\_work}, or \texttt{general}. \\
\texttt{resolved\_by\_paper} & Whether the paper itself appears to resolve the issue. \\
\texttt{in\_related\_work} & Whether the evidence comes from related work or background discussion. \\
\bottomrule
\end{tabular}
\endgroup
\end{table*}

\paragraph{Validation rules.}

Before tuples enter the evidence graph, SGHA applies deterministic validation and cleanup. The relation must belong to the fixed relation vocabulary, and the subject, object, and evidence fields must be present. The supporting evidence span must match the parsed paper text after normalization, so that the tuple remains traceable to the source paper. SGHA removes generic subjects such as ``method,'' ``algorithm,'' or ``our method,'' filters overly generic task labels, drops evaluation tuples that do not include task, dataset, or metric context, and normalizes metadata such as relation polarity. For example, \texttt{fails\_under} and \texttt{limited\_by} are treated as negative relations, while \texttt{assumes} is treated as neutral.

\subsection{Graph Construction and Motif Detection}

Graph construction and motif detection are deterministic. There is no LLM prompt for this stage. SGHA inserts validated paper objects and evidence tuples into a typed evidence graph.

\paragraph{Node and edge types.}

\begin{PromptBox}
Node types:
  Paper
  Method
  Task
  Dataset
  Metric
  Assumption
  Result
  Limitation
  FailureCondition
  Claim
  Gap
  Hypothesis

Edge types:
  mentions
  proposes
  evaluated_on
  uses_dataset
  measured_by
  improves_over
  fails_under
  assumes
  limited_by
  contradicts
  addresses
  not_addressed_by
  supports_gap
  weakens_gap
  has_mechanism
  feasible_with
  derived_from
  counterevidence_for
  partially_addresses_gap
  relaxes_assumption_of
  handles_failure_of
\end{PromptBox}

\paragraph{Motif families.}

SGHA searches the typed graph using predefined motif detectors. A motif hit instantiates a candidate structural gap, but does not by itself verify that the gap is valid or novel.

\begin{PromptBox}
Motif families:
  shared_failure_condition
  repeated_unaddressed_limitation
  conflicting_claims
  sparse_evaluation
  assumption_mismatch
  unrealistic_assumption
  shared_unrealistic_assumption
  theory_practice_gap
  transfer_solution_gap
  unresolved_tradeoff
  missing_stress_test
  claim_without_measurement
  missing_baseline_comparison
\end{PromptBox}

\paragraph{Candidate-gap schema.}

\begin{PromptBox}
CandidateGap:
  gap_id: string
  gap: string
  target: string
  scope: string
  mechanism: string
  supporting_evidence: list
  counterevidence: list
  traceability_path: list
  novelty_score: float
  feasibility_score: float
  impact_score: float
  overall_score: float
  motif_type: string
  motif_id: string
  paper_ids: list[string]
  extraction_uncertainty: float
  counterevidence_edges: list
  counterevidence_papers: list
  counterevidence_status: string
  counterevidence_summary: string
  remaining_gap_scope: string
  novelty_status: string
  novelty_rationale: string
\end{PromptBox}

The graph builder normalizes selected aliases, removes junk method labels, deduplicates parallel edges, and keeps provenance on graph edges. Candidate gaps are created from motif hits, not from arbitrary missing graph edges.

\subsection{Novelty and Corpus-Coverage Filtering}

After motif detection, SGHA applies a corpus-grounded novelty and coverage screen. This screen removes candidates that appear trivial, already solved, or already covered by nearby evidence in the selected corpus. It is not an external literature novelty proof.

\paragraph{Prompt template.}

\begin{PromptBox}
You are a senior ML researcher evaluating structural gaps found by automated analysis of a paper corpus.
Your job is to FILTER OUT weak gaps. Most gaps will be trivial -- be skeptical by default.

Research area: {topic_description}

Classify each gap into exactly one category:

- "trivial": gap is just "method X can fail" or "X has limitations" with no named assumption or failure condition.
- "known_open": specific named gap experts know about but have not solved.
- "known_solved": already substantially addressed in literature published after 2021.
- "novel": non-obvious pattern only visible by reading across multiple papers.

Be strict about trivial. Be fair about known_open when the gap names a specific assumption or condition.

{gap_blocks}

Respond with a JSON object:
{
  "classifications": [
    {"gap_index": 1, "class": "<trivial|known_open|known_solved|novel>", "reason": "<one sentence>"}
  ]
}
\end{PromptBox}

\paragraph{Filtering rules.}

SGHA keeps candidates labeled \texttt{novel}, \texttt{known\_open}, or \texttt{partially\_addressed\_in\_corpus}. It removes candidates labeled \texttt{trivial}, \texttt{known\_solved}, or \texttt{known\_solved\_in\_corpus}. If corpus counterevidence already marks a candidate as solved within the selected evidence base, the LLM classification can be skipped. If a batch classification returns fewer labels than expected, missing labels default to keep with an error rationale; later verification applies the stricter gate.

\subsection{Verification Agents and Hard Gate}

Novelty survivors selected for verification are reviewed by role-specific agents. The agents use the same output schema but different role instructions. The support agent looks for evidence supporting the gap. The skeptic agent looks for counterevidence or already-addressed risk. The feasibility agent checks whether the problem can plausibly be studied. The mechanism agent checks whether there is a plausible explanation for the gap. The verification critic gives a conservative final assessment.

\paragraph{Verification-agent output schema.}

\begin{PromptBox}
{
  "gap_id": "<copy from gap>",
  "agent_name": "<support|skeptic|feasibility|mechanism|critic>",
  "summary": "<one sentence assessment>",
  "evidence": [{"paper_id": "...", "evidence_text": "..."}],
  "counterevidence": [{"paper_id": "...", "evidence_text": "..."}],
  "citations": ["paper_id_1", "paper_id_2"],
  "confidence": 0.0,
  "failure_modes": ["..."],
  "scores": {
    "evidence_support": 0.0,
    "novelty": 0.0,
    "feasibility": 0.0,
    "specificity": 0.0,
    "scope_validity": 0.0
  }
}
\end{PromptBox}

\paragraph{Repair prompt.}

If an agent response is not valid JSON, SGHA retries once with a repair prompt:

\begin{PromptBox}
The previous response was invalid JSON. Return ONLY valid JSON with these keys:
gap_id, agent_name, summary, evidence, counterevidence, citations, confidence, failure_modes, scores.

confidence must be a float between 0.0 and 1.0.
scores must be a dict with keys: evidence_support, novelty, feasibility, specificity, scope_validity (all floats 0.0-1.0).

Gap ID: {gap_id}
Agent: {agent_name}
\end{PromptBox}

\paragraph{Hard verification gate.}

The hard verification gate is deterministic. A reviewed gap enters formulation only if the gate accepts it.

\begin{PromptBox}
verification_gate:
  enabled: true
  mode: survival_score
  min_survival_score: 0.60
  require_all_agents: true
  require_critic_non_reject: true
  min_critic_confidence: 0.50
  fail_on_agent_parse_failure: true
  allow_reviewed_only_fallback: false
\end{PromptBox}

A candidate fails the gate if required agent outputs are missing, if an agent has a disqualifying parse failure, if the survival score is missing or below threshold, or if the critic rejects the candidate or has insufficient confidence.

\subsection{Direct Formulation}

The direct-formulation prompt is used only for gaps that pass the hard verification gate. It asks for a single evidence-close problem formulation.

\paragraph{Prompt template.}

\begin{PromptBox}
You are formulating ONE direct, coherent research problem from a single verification-passed structural gap. This is a PROPOSAL (no results yet).

DECLARED RUN CONTEXT
- domain_name: {domain_name}
- topic_description: {topic_description}
- keyphrases: {keyphrases}

STRICT COHERENCE RULES:
- exactly one core setting, one main assumption/failure/limitation, one core objective
- one plausible theorem OR algorithm OR benchmark/empirical target
- do NOT combine unrelated methods/assumptions/failures into a collage
- do NOT introduce domain-specific concepts unless they appear in the declared run context, source gap, supporting paper titles, extracted evidence, or verification findings
- use the domain's own entities, objectives, assumptions, feedback/measurement model, and evaluation targets
- if the evidence is thin, say so and lower the scores

SOURCE VERIFICATION-PASSED GAP
- gap_id: {gap_id}
- motif_type: {motif_type}
- target: {target}
- gap statement: {gap}
- mechanism: {mechanism}
- scope: {scope}
- supporting paper titles: {papers}
- extracted evidence snippets: {extracted_evidence}

VERIFICATION AGENT FINDINGS:
{verification_summary}
\end{PromptBox}

\paragraph{Output schema.}

\begin{PromptBox}
{
  "direct_title": "",
  "direct_problem_statement": "",
  "proposal_style_abstract": "",
  "core_setting": "",
  "core_assumption_or_failure": "",
  "core_objective": "",
  "possible_theorem_target": "",
  "possible_algorithmic_target": "",
  "possible_empirical_target": "",
  "coherence_score": 0.0,
  "specificity_score": 0.0,
  "feasibility_score": 0.0,
  "term_soup_risk": 0.0,
  "main_risk": "",
  "falsification_condition": "",
  "recommendation": "KEEP | MAYBE | DROP"
}
\end{PromptBox}

SGHA marks a direct formulation as strong when coherence, specificity, and term-soup-risk checks pass and the recommendation is \texttt{KEEP}. Low-coherence or high-term-soup formulations are deprioritized or removed before later stages.

\subsection{Ambition Expansion and Critic}

Ambition expansion starts from a direct formulation. The goal is to broaden the problem where useful while keeping it tied to the verified evidence.

\paragraph{Ambition-expansion prompt template.}

\begin{PromptBox}
You are doing CONTROLLED ambition expansion of ONE clean, single-gap research formulation.
Goal: a genuinely NON-INCREMENTAL but still coherent formulation -- NOT a multi-concept collage,
and NOT just validating one named method in one more setting.

SOURCE DIRECT FORMULATION:
{direct_formulation}

A formulation is INCREMENTAL if its central contribution is only:
- evaluating one named method in one additional setting
- testing robustness of one paper's method without changing the scientific question
- a small benchmark variation
- restating a limitation already identified by one source paper
- depending heavily on one named method
- a cleaner title for the same narrow gap

A formulation is NON-INCREMENTAL if it:
- identifies a broader problem class
- relaxes, removes, or proves necessity of a key assumption
- characterizes a boundary / impossibility / phase transition / identifiability limit / failure regime
- proposes a constructive alternative at method/system/mechanism/evaluation-class level
- defines benchmark/evaluation protocol exposing a general failure mode
- gives a unifying explanation connecting observations under one structural condition
\end{PromptBox}

\paragraph{Ambition-variant schema.}

\begin{PromptBox}
{
  "title": "",
  "problem_statement": "",
  "proposal_style_abstract": "",
  "core_setting": "",
  "generalized_axis": "",
  "core_assumption_or_failure": "",
  "core_objective": "",
  "contribution_type": "theorem|algorithm|lower_bound|benchmark|empirical_study|impossibility_boundary",
  "theorem_target": "",
  "algorithmic_target": "",
  "empirical_target": "",
  "broader_problem_class": "",
  "assumption_shift": "",
  "boundary_or_failure_regime": "",
  "constructive_or_explanatory_target": "",
  "method_class_scope": "",
  "named_method_dependency": "low|medium|high",
  "why_not_just_validation": "",
  "why_not_incremental": "",
  "why_not_term_soup": "",
  "non_incrementality_score": 0.0,
  "incrementality_risk": 0.0,
  "main_risk": "",
  "falsification_condition": "",
  "coherence_score": 0.0,
  "ambition_score": 0.0,
  "specificity_score": 0.0,
  "feasibility_score": 0.0,
  "term_soup_risk": 0.0,
  "recommendation": "KEEP|MAYBE|DROP"
}
\end{PromptBox}

\paragraph{Ambition-critic prompt template.}

\begin{PromptBox}
You are an INDEPENDENT critic. You did NOT write the formulation below and you must NOT rewrite,
fix, or improve it. You only JUDGE whether it is genuinely non-incremental.

Judge strictly:
1. Is this genuinely non-incremental relative to the source direct formulation?
2. Did it change the scientific object, or just wording?
3. Is broader_problem_class real and specific?
4. Is the contribution meaningful beyond one named method?
5. Is the claimed boundary/impossibility/algorithm/benchmark supported by source evidence?
6. Is contribution_type inflated?

SOURCE DIRECT FORMULATION:
{direct_formulation}

VARIANT TO JUDGE:
{ambition_variant}
\end{PromptBox}

\paragraph{Ambition-critic schema.}

\begin{PromptBox}
{
  "critic_non_incrementality_score": 0.0,
  "critic_incrementality_risk": 0.0,
  "critic_fake_ambition_risk": 0.0,
  "critic_named_method_dependency": "low|medium|high",
  "critic_broader_problem_class_valid": true,
  "critic_contribution_type_valid": true,
  "critic_supported_by_source": "strong|moderate|weak",
  "critic_quality_label": "STRONG_NON_INCREMENTAL|VALID_BUT_MODEST|INCREMENTAL_REPHRASING|FAKE_AMBITION|DROP",
  "critic_reason": ""
}
\end{PromptBox}

A variant passes the critic when it is sufficiently non-incremental, source-supported, and specific, with low fake-ambition and incrementality risk. SGHA keeps critic-passing variants for family consolidation.

\subsection{Family Consolidation}

Family consolidation is deterministic in the main runs. It groups critic-passing variants into project families using strict anchors such as shared verified gaps, shared direct formulations, overlapping supporting papers, and strong identity overlap.

\paragraph{Family schema.}

\begin{PromptBox}
{
  "family_id": "family:01",
  "family_title": "",
  "member_variant_ids": [],
  "representative_variant_id": "",
  "representative_title": "",
  "family_problem_statement": "",
  "proposal_style_abstract": "",
  "theorem_target": "",
  "algorithmic_target": "",
  "empirical_target": "",
  "main_risk": "",
  "falsification_condition": "",
  "critic_reason": "",
  "research_object": "",
  "problem_class": "",
  "assumption_shift": "",
  "failure_boundary_or_mechanism": "",
  "constructive_or_evaluation_target": "",
  "source_verified_gaps": [],
  "source_direct_formulations": [],
  "supporting_papers": [],
  "related_seed_papers": [],
  "quality_label": "A_STRONG|B_PROMISING_NEEDS_REFRAMING|C_LOW_PRIORITY|D_DROP",
  "recommended_action": "READ_FIRST|REFRAME|READ_LATER|DROP"
}
\end{PromptBox}

The consolidation stage avoids weak transitive grouping. All selected variants must be assigned consistently, and no invented variant identifiers are allowed.

\subsection{Formal Problem Formulation}

For each final project family, SGHA produces a semi-formal problem object. The prompt asks for a structured problem statement rather than a finished theorem.

\paragraph{Prompt template.}

\begin{PromptBox}
You are SGHA's domain-general formal problem formulation stage.

You must formalize ONE existing project family. Do NOT invent a new research direction, new family,
or new claim of results. Use only the source family and formulation evidence below.

STRICT RULES:
- Formalize the existing project family only.
- Prefer clear semi-formal structure over fake mathematics.
- If a term is vague, define it cautiously or put it in ambiguity_flags.
- If the feedback/data/measurement model is unclear, say so.
- If the objective cannot be formalized from evidence, set formalization_confidence to "low".
- Every symbol introduced in the formal_problem_statement must appear in the variable table.
- Do NOT use metadata variables such as R = research object, C = problem class,
  A = assumption shift, or B = boundary.
- Variables must correspond to problem-level objects: actions, observations, latent factors,
  measurements, rewards/outcomes, policies/decisions, constraints, objectives, estimators,
  environments, systems, functions, processes, or distributions.
- Every major assumption must be marked kept, relaxed, removed, or questioned.
- Do not say "we prove", "we show", or imply results already exist.

PROJECT FAMILY:
{project_family}
\end{PromptBox}

\paragraph{Formal-problem schema.}

\begin{PromptBox}
{
  "plain_language_problem": "",
  "formal_problem_statement": "",
  "mathematical_setup": {
    "entities": [""],
    "variables": [
      {
        "symbol": "",
        "meaning": "",
        "type": "scalar | vector | set | distribution | process | function | other",
        "source": "from evidence | introduced for formalization"
      }
    ],
    "data_or_observations": "",
    "feedback_or_measurement_model": "",
    "decision_variables_or_outputs": "",
    "objective": "",
    "constraints": "",
    "success_criterion": ""
  },
  "assumptions": [
    {
      "name": "",
      "description": "",
      "status": "kept | relaxed | removed | questioned",
      "source_evidence": [""]
    }
  ],
  "open_question": "",
  "possible_result_types": {
    "theorem": "",
    "algorithm": "",
    "empirical_or_benchmark": ""
  },
  "evaluation_protocol": "",
  "ambiguity_flags": [
    {
      "term": "",
      "why_ambiguous": "",
      "what_user_must_define": ""
    }
  ],
  "source_grounding": {
    "source_verified_gaps": [""],
    "supporting_papers": [""],
    "representative_formulation": "",
    "critic_reason": ""
  },
  "formalization_confidence": "high | medium | low",
  "formalization_risk": "",
  "requires_human_definition": true
}
\end{PromptBox}

If the formal-problem output is not valid JSON, SGHA retries once with a repair prompt. If repair fails, the system uses a source-grounded low-confidence fallback and marks the formalization as requiring human definition.

\subsection{Profile-Conditioned Additions}

Profile-conditioned SGHA uses the same downstream prompts and gates after corpus construction. The profile changes the evidence base and prioritization, not the validity checks.

\paragraph{Seed-paper schema.}

\begin{PromptBox}
SeedPaper:
  seed_paper_id
  title
  authors
  year
  venue
  abstract
  local_pdf_path
  arxiv_id
  openreview_id
  doi
  url
  source
  relation_to_seed_profile
  is_manual_seed_paper
  seed_label
  metadata_incomplete
  provenance
\end{PromptBox}

\paragraph{Profile schema.}

\begin{PromptBox}
{
  "seed_label": "",
  "topic": "",
  "seed_titles": [],
  "keyphrases": [],
  "method_terms": [],
  "task_terms": [],
  "assumption_terms": [],
  "benchmark_terms": [],
  "generated_openreview_queries": [],
  "venues": [],
  "exclusion_terms": [],
  "profile_summary": ""
}
\end{PromptBox}

In profile-conditioned mode, Stage 7 and later stages may receive seed-alignment information and related seed papers as context. Verification, formulation, ambition criticism, family consolidation, and formalization remain the same.

\subsection{Evolutionary Exploration}

The evolutionary branch is separate from the main family-report path. It starts from literature-grounded seeds and generates nearby variants by changing parts of the problem representation, such as assumptions, mechanisms, objectives, settings, or regimes. These variants are critiqued, selected, and refined over multiple rounds.

\paragraph{Evolutionary-candidate schema.}

\begin{PromptBox}
EvolutionaryHypothesis:
  candidate_id: string
  source_seed_id: string
  parent_ids: list[string]
  generation: integer
  mutation_type: string
  title: string
  problem_statement: string
  gap: string
  target: string
  scope: string
  mechanism: string
  proposal_style_abstract: string
  source_grounding: list
  motif_lineage: list
  evolution_score: float
  rank: integer
  selected: boolean
\end{PromptBox}

The evolutionary branch is used for breadth. Its selected or ranked hypotheses can be evaluated with the same formulation-only rubric, but they are reported separately from the final verification-gated project families.

\subsection{Final Report Auditing}

Final report rendering is deterministic in the main runs. The audit checks that final families are present or that a low-signal status is recorded; that direct formulations trace to verification-passed gaps; that selected variants trace to direct formulations; that final families trace to accepted variants; and that formal problem JSON exists when formalization is enabled. It also checks that novelty-only or reviewed-only candidates do not enter the default family-report path.

The final status is one of:

\begin{PromptBox}
PIPELINE_COMPLETED_PASS
PIPELINE_COMPLETED_LOW_SIGNAL
PIPELINE_COMPLETED_AUDIT_FAILURE
\end{PromptBox}

A low-signal status means the pipeline completed but did not produce a final family because the evidence or downstream checks did not support one.

%% file: appendix/appendixb.tex
\section{LLM-Judge Rubric and Candidate Normalization}
\label{app:judge-details}

In this appendix, we describe how generated candidates are normalized and evaluated by the formulation-only LLM judges.

\subsection{Candidate Packet Format}

Before judging, outputs from all methods are converted into a common candidate-packet format. The packet preserves the fields produced by the original method. Fields that are not produced by a method are left as \texttt{not provided}; they are not filled in manually after generation.

\begin{PromptBox}
CandidatePacket:
  candidate_id: string
  method_label: hidden during judging
  domain: string
  title: string
  problem_statement: string
  motivation_or_abstract: string
  proposed_direction: string
  expected_contribution: string
  evaluation_plan: string
  risks_or_caveats: string
  source_context_or_grounding: string
  assumptions_or_problem_setup: string
  formal_problem_statement: string
  ambiguity_or_missing_definitions: string
\end{PromptBox}

This format lets the judge compare artifacts produced by different systems while preserving their natural structure. SGHA final families often include formal problem statements, assumptions, source provenance, and ambiguity flags. However, ideation baselines often include motivation, proposed directions, evaluation plans, and caveats. The normalization step keeps these differences visible.

\subsection{Formulation-Only Judge Prompt}

The formulation-only judge evaluates the quality of a research-problem formulation. 

\paragraph{Judge prompt template.}

\begin{PromptBox}
You are a strict, skeptical senior research reviewer evaluating research-problem formulation quality only.

Your task is to judge the formulation as a problem statement, not as a project plan.
Do not evaluate implementation plans, experiments, software engineering details, or actionability except where they clarify the formulation itself.
Do not reward a candidate for having a polished plan if the underlying research problem is vague or poorly posed.

Judge only the provided blinded candidate text.
Do not use external knowledge.
Do not perform an external novelty check.
If a field says "not provided", treat it as missing.

Score the candidate using the rubric below.
Return only valid JSON matching the requested schema.

CANDIDATE PACKET:
{candidate_packet}

RUBRIC:
{rubric}

SCORE ANCHORS:
{score_anchors}

CAP RULES:
{cap_rules}

Return JSON only.
\end{PromptBox}

\subsection{Formulation Rubric}

Table~\ref{tab:app-formulation-rubric} lists the ten formulation-only criteria. Each criterion is scored from 0 to 10.

\begin{table*}[t]
\centering
\small
\caption{Formulation-only LLM-judge rubric.}
\label{tab:app-formulation-rubric}
\begingroup
\renewcommand{\arraystretch}{1.12}
\begin{tabular}{L{0.28\linewidth}L{0.62\linewidth}}
\toprule
\tableheader
\theadcell{Criterion} & \theadcell{What the judge evaluates} \\
\midrule
\textbf{Problem-definition clarity} & Whether the candidate states a clear research problem rather than a broad topic or loose motivation. \\
\textbf{Technical specificity} & Whether the formulation names concrete objects, settings, mechanisms, assumptions, metrics, or target results. \\
\textbf{Well-posedness} & Whether the problem has enough structure to be studied, refined, or formalized. \\
\textbf{Assumption-boundary clarity} & Whether the formulation makes clear which assumptions are used, relaxed, questioned, removed, or missing. \\
\textbf{Formalizability} & Whether the problem can plausibly be written as a theorem, algorithmic objective, benchmark protocol, impossibility result, or other formal research target. \\
\textbf{Nontriviality} & Whether the candidate goes beyond a simple application, small robustness check, or generic ``apply X to Y'' idea. \\
\textbf{Scope control} & Whether the problem is focused enough to be studied rather than combining many loosely related concepts. \\
\textbf{Source-grounded specificity} & Whether the formulation is tied to provided source context, evidence, or literature-derived details. \\
\textbf{Ambiguity hygiene} & Whether the candidate explicitly states missing definitions, unclear feedback models, unresolved assumptions, or caveats. \\
\textbf{Overall formulation quality} & The judge's overall assessment of the research-problem formulation under the rubric. \\
\bottomrule
\end{tabular}
\endgroup
\end{table*}

\subsection{Score Anchors}

The judges use the anchors in Table~\ref{tab:app-score-anchors}. These anchors are included in the judge prompt to make the scale consistent across models.

\begin{table}[t]
\centering
\small
\caption{Score anchors used by the formulation-only judges.}
\label{tab:app-score-anchors}
\begingroup
\renewcommand{\arraystretch}{1.12}
\begin{tabular}{C{0.12\linewidth}L{0.74\linewidth}}
\toprule
\tableheader
\theadcell{Score} & \theadcell{Interpretation} \\
\midrule
\textbf{0} & No usable formulation. \\
\textbf{1} & Mostly incoherent or unrelated to a research problem. \\
\textbf{2} & Topic-level idea with almost no problem structure. \\
\textbf{3} & Very vague formulation with major missing pieces. \\
\textbf{4} & Weak formulation; some direction is visible, but the problem is poorly specified. \\
\textbf{5} & Plausible idea, but important assumptions, scope, or definitions are missing. \\
\textbf{6} & Plausible formulation with useful structure, but still needing substantial refinement. \\
\textbf{7} & Strong formulation with clear problem structure and reasonable technical specificity. \\
\textbf{8} & Very strong formulation; well posed, grounded, and mostly ready for expert refinement. \\
\textbf{9} & Excellent formulation with unusually clear assumptions, scope, and formal target. \\
\textbf{10} & Exceptional formulation; clear, grounded, formalizable, and close to research-ready. \\
\bottomrule
\end{tabular}
\endgroup
\end{table}

\subsection{Score Consistency Rules}

The judge prompt includes cap rules to keep scores consistent when important information is missing. These rules do not assign scores directly. They limit the maximum score for a criterion when the candidate lacks the corresponding field or structure.

\begin{table*}[t]
\centering
\small
\caption{Score consistency rules used by the formulation-only judge.}
\label{tab:app-cap-rules}
\begingroup
\renewcommand{\arraystretch}{1.12}
\begin{tabular}{L{0.36\linewidth}L{0.52\linewidth}}
\toprule
\tableheader
\theadcell{Condition} & \theadcell{Rule} \\
\midrule
Problem statement is vague or mostly a topic description & Overall formulation quality should be at most 6. \\
Formal problem statement is not provided & Well-posedness and formalizability should be at most 6. \\
Assumptions or setup are not provided & Well-posedness and assumption-boundary clarity should be at most 6. \\
Ambiguity or missing-definition field is not provided & Ambiguity hygiene should be at most 6. \\
Source or context grounding is not provided & Source-grounded specificity should be at most 4. \\
Candidate combines many loosely related concepts & Scope control should be at most 4. \\
Candidate is mainly ``apply X to Y'' or one-method validation & Nontriviality should be at most 6. \\
No clear formal skeleton is visible & Overall formulation quality should be at most 7. \\
\bottomrule
\end{tabular}
\endgroup
\end{table*}

\subsection{Judge Response Schema}

Each judge returns a strict JSON object. The schema records the criterion-level scores, a recommendation, confidence, strengths, weaknesses, rationale, and a novelty caveat.

\begin{PromptBox}
{
  "candidate_id": "",
  "domain": "",
  "scores": {
    "problem_definition_clarity_10": 0,
    "technical_specificity_10": 0,
    "well_posedness_10": 0,
    "assumption_boundary_clarity_10": 0,
    "formalizability_10": 0,
    "nontriviality_10": 0,
    "scope_control_10": 0,
    "source_grounded_specificity_10": 0,
    "ambiguity_hygiene_10": 0,
    "overall_formulation_quality_10": 0
  },
  "recommended_action": "READ_FIRST | PROMISING_NEEDS_REFINEMENT | NEEDS_REFRAMING | DROP_OR_DEPRIORITIZE",
  "confidence": "LOW | MEDIUM | HIGH",
  "strengths": ["..."],
  "weaknesses": ["..."],
  "rationale": "...",
  "novelty_caveat": "Novelty potential judged only from provided text; no external novelty check performed."
}
\end{PromptBox}

The reported \texttt{overall\_formulation\_quality\_10} score is the judge's holistic assessment under the rubric. It is not computed as a weighted average of the other criteria.

\subsection{Judge Panel and Blinding}

We use five LLM judges from different providers:

\begin{PromptBox}
anthropic/claude-sonnet-4
openai/gpt-5.6-sol-pro
x-ai/grok-4.5
moonshotai/kimi-k3
google/gemini-3.6-flash
\end{PromptBox}

Each judge receives the same blinded candidate packet format, rubric, score anchors, cap rules, and JSON response schema. Method labels are hidden during scoring. The blinding key is used only after scoring, during postprocessing. The judge panel is used only after all candidate outputs have been frozen.

\subsection{Calibration Examples}

Before scoring the real candidates, each judge is checked on calibration examples. These examples are designed to test whether the judge applies the rubric in the intended direction. The calibration set includes vague topic-level ideas, underformalized formulations, over-broad or term-heavy formulations, and stronger source-grounded formal problem statements.

Calibration is used as a quality check for the judge behavior. It is not used to tune separate prompts for different judges; the rubric and response schema remain fixed.

\subsection{Personalized Judge}

Profile-conditioned candidates are evaluated with a personalized judge rubric. This rubric keeps the ten formulation-quality criteria and adds profile-specific criteria. The personalized judge uses only the provided profile context and candidate text; it does not use external knowledge about the researcher.

\paragraph{Personalized judge prompt template.}

\begin{PromptBox}
You are a strict, skeptical senior research reviewer evaluating personalized research-problem formulations.

Your task is to judge both formulation quality and personalization quality.
Judge profile alignment only from the provided profile context and the candidate text.
Do not use external knowledge about the researcher.
If a field says "not provided", treat it as missing.
Do not reward name-dropping.
A good personalized problem connects to artifact-supported profile themes, source/corpus evidence,
and profile-relevant technical style while formulating a meaningful next problem.
Penalize generic problems that could apply to many researchers in the area.
Penalize off-profile problems even if technically coherent.

PROFILE CONTEXT:
{profile_context}

CANDIDATE PACKET:
{candidate_packet}

Return JSON only.
\end{PromptBox}

\paragraph{Additional personalized criteria.}

\begin{table*}[t]
\centering
\small
\caption{Additional criteria used by the personalized judge.}
\label{tab:app-personalized-criteria}
\begingroup
\renewcommand{\arraystretch}{1.12}
\begin{tabular}{L{0.28\linewidth}L{0.62\linewidth}}
\toprule
\tableheader
\theadcell{Criterion} & \theadcell{Meaning} \\
\midrule
\textbf{Profile alignment} & Whether the candidate problem is supported by the provided profile-derived literature context. \\
\textbf{Profile specificity} & Whether the problem is specific to that context rather than a generic domain problem. \\
\textbf{Intellectual-style match} & Whether the problem matches the profile-associated technical style or research pattern. \\
\textbf{Profile novelty fit} & Whether the candidate plausibly extends the profile-associated literature rather than repeating it. \\
\textbf{Personalization overall} & Overall quality of the personalization, considering the above criteria. \\
\bottomrule
\end{tabular}
\endgroup
\end{table*}

\paragraph{Personalized response schema.}

\begin{PromptBox}
{
  "candidate_id": "",
  "profile": "",
  "scores": {
    "problem_definition_clarity_10": 0,
    "technical_specificity_10": 0,
    "well_posedness_10": 0,
    "assumption_boundary_clarity_10": 0,
    "formalizability_10": 0,
    "nontriviality_10": 0,
    "scope_control_10": 0,
    "source_grounded_specificity_10": 0,
    "ambiguity_hygiene_10": 0,
    "overall_formulation_quality_10": 0,
    "profile_alignment_10": 0,
    "profile_specificity_10": 0,
    "intellectual_style_match_10": 0,
    "profile_novelty_fit_10": 0,
    "personalization_overall_10": 0
  },
  "recommended_action": "READ_FIRST | PROMISING_NEEDS_REFINEMENT | NEEDS_REFRAMING | DROP_OR_DEPRIORITIZE",
  "confidence": "LOW | MEDIUM | HIGH",
  "strengths": ["..."],
  "weaknesses": ["..."],
  "rationale": "...",
  "personalization_caveat": "Profile alignment judged only from provided profile context; no external profile knowledge used.",
  "novelty_caveat": "Novelty potential judged only from provided text; no external novelty check performed."
}
\end{PromptBox}

The personalized judge also uses additional consistency rules. For example, candidates that only name-drop the researcher receive low profile-alignment scores, and generic domain problems without profile-specific evidence receive lower profile-specificity scores.

%% file: appendix/appendixc.tex
\section{Additional Result Tables}
\label{app:additional-results}

This appendix provides detailed count and score tables that support the main results. The main text reports the central trends; here we include the full SGHA stage counts and selected per-domain formulation scores.

\subsection{Full SGHA Pipeline Counts}

Table~\ref{tab:app-full-pipeline-counts} reports the full SGHA count funnel for the five main domains. The table separates paper-level counts, graph-level counts, gap-level counts, and formulation-level counts. These quantities are not one-to-one: one paper can produce many evidence tuples, several tuples can instantiate one motif, and several accepted variants can later be consolidated into one project family.

\begin{table*}[t]
\centering
\scriptsize
\caption{Full SGHA pipeline counts across the five main domains. Motif hits are deterministic graph-pattern matches. Novelty survivors are candidate gaps retained after corpus screening. Reviewed gaps are sent to verification agents. Verified gaps pass the hard verification gate. Accepted variants are ambition-expanded formulations that pass the independent critic.}
\label{tab:app-full-pipeline-counts}
\begingroup
\renewcommand{\arraystretch}{1.12}
\setlength{\tabcolsep}{3pt}
\begin{tabular}{L{0.21\linewidth}rrrrrrrrrrrrrr}
\toprule
\tableheader
\theadcell{Domain} &
\theadcell{Selected} &
\theadcell{Parsed} &
\theadcell{Extracted} &
\theadcell{Tuples} &
\theadcell{Nodes} &
\theadcell{Edges} &
\theadcell{Motif\\hits} &
\theadcell{Novelty\\surv.} &
\theadcell{Reviewed} &
\theadcell{Verified} &
\theadcell{Direct} &
\theadcell{Accepted\\variants} &
\theadcell{Families} &
\theadcell{Formal} \\
\midrule
Bandits & 250 & 221 & 221 & 1554 & 9214 & 9776 & 173 & 30 & 13 & 6 & 6 & 4 & 3 & 3 \\
In-context learning & 250 & 193 & 193 & 1559 & 8608 & 9083 & 147 & 21 & 16 & 12 & 12 & 4 & 4 & 4 \\
Reasoning / test-time computation & 250 & 137 & 137 & 1139 & 6116 & 6535 & 105 & 29 & 12 & 7 & 7 & 1 & 1 & 1 \\
Offline reinforcement learning & 250 & 248 & 245 & 2176 & 10600 & 12045 & 193 & 8 & 8 & 2 & 2 & 1 & 1 & 1 \\
Uncertainty estimation & 250 & 250 & 248 & 2206 & 11097 & 12055 & 218 & 17 & 15 & 12 & 12 & 9 & 6 & 6 \\
\midrule
\tabtotal
\textbf{Total} & \textbf{1250} & \textbf{1049} & \textbf{1044} & \textbf{8634} & \textbf{45635} & \textbf{49494} & \textbf{836} & \textbf{105} & \textbf{64} & \textbf{39} & \textbf{39} & \textbf{19} & \textbf{15} & \textbf{15} \\
\bottomrule
\end{tabular}
\endgroup
\end{table*}

\subsection{Selected Per-Domain Formulation Scores}

Table~\ref{tab:app-per-domain-formulation-scores} reports per-domain formulation scores for the main methods. The main text reports method-level averages. Here, we show the overall score and three criteria that are especially relevant to SGHA: source grounding, formalizability, and ambiguity hygiene.

\begin{table*}[t]
\centering
\scriptsize
\caption{Selected per-domain formulation-quality scores. Scores are averaged over retained candidates and LLM judges.}
\label{tab:app-per-domain-formulation-scores}
\begingroup
\renewcommand{\arraystretch}{1.12}
\setlength{\tabcolsep}{4pt}
\begin{tabular}{L{0.25\linewidth}L{0.27\linewidth}rrrr}
\toprule
\tableheader
\theadcell{Domain} &
\theadcell{Method} &
\theadcell{Overall} &
\theadcell{Source\\grounding} &
\theadcell{Formalizability} &
\theadcell{Ambiguity\\hygiene} \\
\midrule
Bandits & SGHA & 5.93 & 7.33 & 5.20 & 7.80 \\
Bandits & AI-Scientist-v2 + Qwen & 4.53 & 5.20 & 4.00 & 2.93 \\
Bandits & AI-Scientist-v2 + Claude Opus & 5.87 & 2.80 & 5.07 & 3.47 \\
Bandits & MOOSE-Star & 1.93 & 3.47 & 1.27 & 1.40 \\
\midrule
In-context learning & SGHA & 5.80 & 7.25 & 5.35 & 7.55 \\
In-context learning & AI-Scientist-v2 + Qwen & 4.35 & 4.60 & 3.40 & 2.60 \\
In-context learning & AI-Scientist-v2 + Claude Opus & 5.80 & 3.00 & 5.30 & 3.70 \\
In-context learning & MOOSE-Star & 1.95 & 4.00 & 1.55 & 1.50 \\
\midrule
Reasoning / test-time computation & SGHA & 7.20 & 7.80 & 6.40 & 7.80 \\
Reasoning / test-time computation & AI-Scientist-v2 + Qwen & 4.60 & 5.40 & 3.40 & 2.40 \\
Reasoning / test-time computation & AI-Scientist-v2 + Claude Opus & 6.00 & 3.40 & 5.40 & 3.40 \\
Reasoning / test-time computation & MOOSE-Star & 2.00 & 4.00 & 1.40 & 1.40 \\
\midrule
Offline reinforcement learning & SGHA & 6.40 & 7.60 & 6.00 & 7.80 \\
Offline reinforcement learning & AI-Scientist-v2 + Qwen & 4.80 & 5.60 & 3.80 & 3.00 \\
Offline reinforcement learning & AI-Scientist-v2 + Claude Opus & 5.40 & 2.80 & 4.80 & 3.00 \\
Offline reinforcement learning & MOOSE-Star & 2.40 & 4.40 & 2.00 & 1.60 \\
\midrule
Uncertainty estimation & SGHA & 5.87 & 7.50 & 5.53 & 7.50 \\
Uncertainty estimation & AI-Scientist-v2 + Qwen & 4.63 & 5.27 & 3.73 & 2.93 \\
Uncertainty estimation & AI-Scientist-v2 + Claude Opus & 5.90 & 3.00 & 5.27 & 3.67 \\
Uncertainty estimation & MOOSE-Star & 2.00 & 3.57 & 1.53 & 1.50 \\
\bottomrule
\end{tabular}
\endgroup
\end{table*}

The per-domain scores are consistent with the aggregate results in the main text: SGHA is strongest on source grounding and ambiguity hygiene across domains, while the strongest ideation baseline is often competitive on overall formulation quality.